\documentclass[journal]{IEEEtran}

\usepackage{cite}

\usepackage[pdftex]{graphicx}
\graphicspath{{./}}
\DeclareGraphicsExtensions{.jpg}

\usepackage{amsmath}
\usepackage{amssymb}
\usepackage{array}
\usepackage{url}

\usepackage[dvipsnames,svgnames,x11names]{xcolor}
\usepackage{hyperref}
\usepackage[utf8]{inputenc}
\usepackage{multirow}
\usepackage{comment}

\usepackage{pifont}
\usepackage{amsmath}
\usepackage{algorithm, algpseudocode}

\def\eg{\emph{e.g.}} 
\def\ie{\emph{i.e.}} 
\def\etal{\emph{et~al.}} 

\newcommand{\zznote}[1]{\textcolor{black}{#1}}

\newlength\savewidth\newcommand\shline{\noalign{\global\savewidth\arrayrulewidth
  \global\arrayrulewidth 1pt}\hline\noalign{\global\arrayrulewidth\savewidth}}
  
\begin{document}
\title{VehicleNet: Learning Robust Visual Representation \\ 
for Vehicle Re-identification}
\author{Zhedong Zheng, Tao Ruan, Yunchao Wei, Yi Yang, Tao Mei
\thanks{Zhedong Zheng, Yunchao Wei and Yi Yang are with the Australian Artificial Intelligence Institute (AAII), University of Technology Sydney, NSW 2007, Australia. E-mail: zhedong.zheng@student.uts.edu.au, yunchao.wei@uts.edu.au, yi.yang@uts.edu.au. Tao Ruan is with the Institute of Information Science at Beijing Jiaotong University, and the Beijing Key Laboratory of Advanced Information Science and Network Technology, Beijing 100044, China. E-mail: 16112064@bjtu.edu.cn. Tao Mei is with AI Research of JD.COM, Beijing 100105, China. E-mail: tmei@live.com. 
Yi Yang is the corresponding author.}
}

\markboth{Journal of \LaTeX\ Class Files,~Vol.~14, No.~8, August~2015}%
{Shell \MakeLowercase{\textit{et al.}}: Bare Demo of IEEEtran.cls for IEEE Journals}

\maketitle

\begin{abstract}

%Vehicle re-identification (re-id) faces significant intra-class variations across different cameras, which demands robust and discriminative image representation.
One fundamental challenge of vehicle re-identification (re-id) is to learn robust and discriminative visual representation, given the significant intra-class vehicle variations across different camera views.
%However, the existing vehicle datasets are usually limited in the number of  training images or contain limited viewpoints, which could compromise the model learning.
As the existing vehicle datasets are limited in terms of training images and viewpoints, we propose to build a unique large-scale vehicle dataset (called VehicleNet) by harnessing four public vehicle datasets, and design a simple yet effective two-stage progressive approach to learning more robust visual representation from VehicleNet.
The first stage of our approach is to learn the generic representation for all domains (i.e., source vehicle datasets) by training with the conventional classification loss. This stage relaxes the full alignment between the training and testing domains, as it is agnostic to the target vehicle domain. The second stage is to fine-tune the trained model purely based on the target vehicle set, by minimizing the distribution discrepancy between our VehicleNet and any target domain. We discuss our proposed multi-source dataset VehicleNet and evaluate the effectiveness of the two-stage progressive representation learning through extensive experiments.
%The first stage is to train the model with the conventional classification loss to learn the generic representation for all domains. Note that the first stage is agnostic to the target vehicle domain, hence do not bring the training domain and the test domain into full alignment. 
%The second stage is to fine-tune the trained model only on the target vehicle set, minimizing the distribution discrepancy between VehicleNet and the target environment. 
%We discuss and analyze the effectiveness of the proposed multi-source dataset VehicleNet and the two-stage progressive learning approach via extensive experiments.
We achieve the state-of-art accuracy of $86.07\%$ mAP on the private test set of AICity Challenge, and competitive results on two other public vehicle re-id datasets, \ie, VeRi-776 and VehicleID. %Furthermore, we provide extensive experiments of developing a robust vehicle re-identification system, and 
We hope this new VehicleNet dataset and the learned robust representations can pave the way for vehicle re-id in the real-world environments. 
%We hope all these efforts could ease the future study for developing a robust vehicle re-identification system.
\end{abstract}


\IEEEpeerreviewmaketitle

\section{Introduction}
\IEEEPARstart{V}{ehicle} re-identification (re-id) is to spot the car of interest in different cameras and is usually viewed as a sub-task of image retrieval problem \cite{zheng2019joint}. It could be applied to the public place for the traffic analysis, which facilitates the traffic jam management and the flow optimization \cite{tang@cityflow}. Yet vehicle re-id remains challenging since it inherently contains multiple intra-class variants, such as viewpoints, illumination and occlusion. 
Thus, vehicle re-id system demands a robust and discriminative visual representation given that the realistic scenarios are diverse and complicated. 
Recent years, Convolutional Neural Network (CNN) has achieved the state-of-the-art performance in many computer vision tasks, including person re-id \cite{sun2017beyond,sun2017svdnet,zheng2018pedestrian} and vehicle re-id \cite{liu2016deep,zhou2018aware,wang2017orientation}, but CNN is data-hungry and prone to over-fitting small-scale datasets. Since the paucity of vehicle training images compromises the learning of robust features, vehicle re-id for the small datasets turn into a challenging problem.

%First of all, the challenge of training data limitation remains. 
One straightforward approach is to annotate more data and re-train the CNN-based model on the augmented dataset. However, it is usually unaffordable due to the annotation difficulty and the time cost. Considering that many vehicle datasets collected in lab environments are publicly available, an interesting problem arises: Can we leverage the public vehicle image datasets to learn the robust vehicle representation? Given vehicle datasets are related and vehicles share the similar structure, more data from different sources could help the model to learn the common knowledge of vehicles. Inspired by the success of large-scale datasets, \eg, ImageNet \cite{deng2009imagenet}, we collect a large-scale vehicle dataset, called VehicleNet.

Intuitively, we could utilize VehicleNet to learn the relevance between different vehicle re-id datasets. Then the robust features could be obtained by minimizing the objective function. However, different datasets are collected in different environments, and contains different biases. Some datasets, such as CompCar \cite{yang2015large}, are mostly collected in the car exhibitions, while other datasets, \eg, City-Flow \cite{tang@cityflow} and VeRi-776 \cite{liu2016deep}, are collected in the real traffic scenarios. Thus, another scientific problem of how to leverage the multi-source vehicle dataset occurs. In several existing works, some researchers resort to transfer learning \cite{pan2009survey}, which aims at transferring the useful knowledge from the labeled source domain to the unlabeled target domain and minimizing the discrepancy between the source domain and the target domain. Inspired by the spirit of transfer learning, in this work, we propose a simple two-stage progressive learning strategy to learn from VehicleNet and adapt the trained model to the realistic environment.

\begin{figure}[t]
\begin{center}
   \includegraphics[width=1\linewidth]{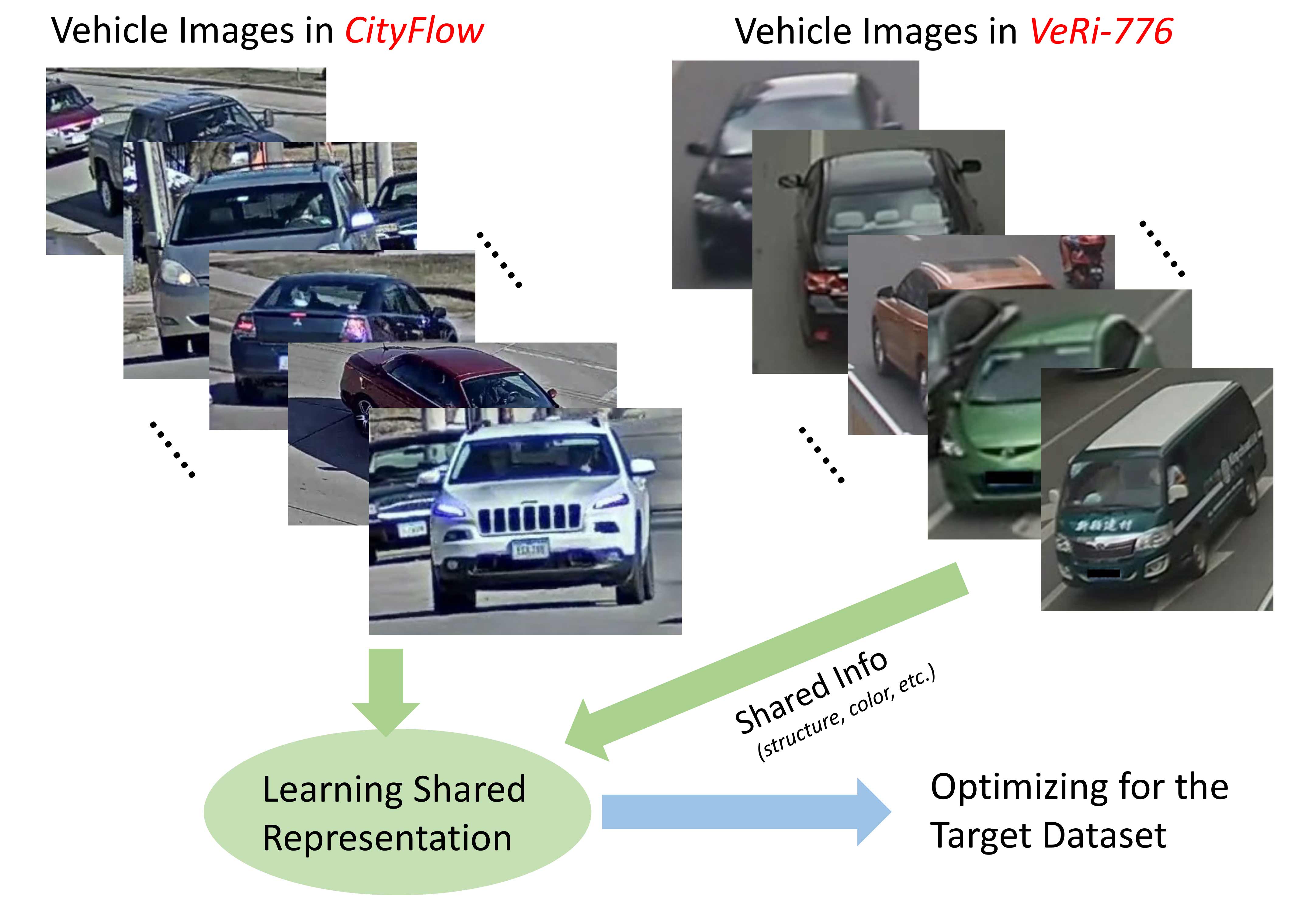}
\end{center}
\vspace{-.2in}
   \caption{The motivation of our vehicle re-identification method by leveraging public datasets. The common knowledge of discriminating different vehicles could be transferred to the final model.}
\label{fig:motivation}
\vspace{-.1in}
\end{figure}

In a summary, to address the above-mentioned challenges, \ie, the data limitation and the usage of multi-source dataset, we propose to build a large-scale dataset, called VehicleNet, via the public datasets and learn the common knowledge of the vehicle representation via two-stage progressive learning (see Figure \ref{fig:motivation}). 
Specifically, instead of only using the original training dataset, we first collect free vehicle images from the web. Comparing with the training set of the CityFlow dataset, we scale up the number of training images from $26,803$ to $434,440$ as a new dataset called VehicleNet. We train the CNN-based model to identify different vehicles, and extract features. With the proposed two-stage progressive learning, the model is further fine-tuned to adapt to the target data distribution, yielding the performance boost. 
In the experiment, we show that it is feasible to train models with a combination of multiple datasets. When training the model with more samples, we observe a consistent performance boost, which is consistent with the observation in some recent works \cite{zheng2019joint,krause2016unreasonable,mahajan2018exploring}. 
Without explicit vehicle part matching or attribute recognition, the CNN-based model learns the viewpoint-invariant feature by ``seeing'' more vehicles.  Albeit simple, the proposed method achieves mAP $75.60\%$ on the private testing set of CityFlow~\cite{tang@cityflow} without extra information. With the temporal and spatial annotation, our method further arrives the $86.07\%$ mAP. The result surpasses the AICity Challenge  champion, who also uses the temporal and spatial annotation.
In a nutshell, our contributions are two-folds:
\begin{itemize}
    \item To address the data limitation, we introduce one large-scale dataset, called VehicleNet, to borrow the strength of the public vehicle datasets, which facilitate the learning of robust vehicle features. In the experiment, we verify the feasibility and effectiveness of learning from VehicleNet.
    \item To leverage the multi-source vehicle images in VehicleNet, we propose a simple yet  effective learning strategy, \ie, the two-stage progressive learning approach. We discuss and analyze the effectiveness of the two-stage progressive learning approach. The proposed method has achieved competitive performance on the CityFlow benchmark as well as two public vehicle re-identification datasets, \ie, VeRi-776~\cite{liu2016deep} and VehicleID~\cite{liu2016pku}.
\end{itemize}
%The rest of this paper is organized as follows. Section \ref{sec:relatedwork} reviews and discusses the related works. In Section \ref{sec:datasetcollection}, we illustrate the vehicle re-id dataset and the task definition, followed by the proposed two-stage progressive learning in Section \ref{sec:method}. Extensive experiments and ablation studies are in Section \ref{sec:experiment}, and the conclusion is draw in Section \ref{sec:conclusion}.

\section{Related Work} \label{sec:relatedwork}
\subsection{Vehicle Re-identification}
Vehicle re-identification (re-id) demands robust and discriminative image representation. The recent progress of vehicle re-identification has been due to two aspects: 1) the availability of the new vehicle datasets \cite{tang@cityflow,liu2016deep,liu2016pku,yan2017exploiting} and 2) the discriminative vehicle feature from deeply-learned models.
%Vehicle re-identification demands a robust and discriminative feature.
Zapletal \etal \cite{zapletal2016vehicle} first collect a large-scale dataset with vehicle pairs and extract the color histograms and oriented gradient histograms feature to discriminate different cars. With recent advance in Convolutional Neural Network (CNN), Liu \etal \cite{liu2016large} combine the CNN-based feature with the traditional hand-crafted features to obtain the robust feature. \zznote{Qian \etal \cite{qian2017multi} and Guo \etal \cite{guo2018learning} propose to aggregate the multi-level feature to enrich the representation.} To take fully advantages of the fine-grained patterns, Wang \etal  \cite{wang2017orientation} first explore the vehicle structure and then extract the part-based CNN features according to the location of key points. Besides, Shen \etal \cite{shen2017learning} involve the temporal-spatial information into the model training as well as the inference process. Another line of works regards vehicle re-identification as a metric learning problem, and explore the objective functions to help the representation learning. Triplet loss has been widely studied in person re-id \cite{hermans2017defense,zheng2017dual,ding2020adaptive}, and also has achieved successes in the vehicle re-id \cite{liu2016deep}. Zhang \etal \cite{zhang2017improving} further company the classification loss with triplet loss, which further improves the re-identification ability.
Furthermore, Yan \etal \cite{yan2017exploiting} propose a multi-grain ranking loss to discriminate the appearance-similar cars. Besides, some works also show the attributes, \eg, color, manufactories and wheel patterns, could help the model to learn the discriminative feature \cite{lin2017improving,tang@cityflow,wang2018transferable}.

\subsection{Dataset Augmentation}
Many existing works focus on involving more samples to boost the training. One line of works leverage the generative model to synthesize more samples for training. Wu \etal~\cite{wu2019ace} and Yue \etal~\cite{yue2019domain} propose to transfer the image into different image styles, \eg, weather conditions, and learn the robust feature for semantic segmentation.
In a similar spirit, Zheng \etal \cite{zheng2017unlabeled,zheng2019joint} utilize the Generative Adversarial Network (GAN) \cite{goodfellow2014generative} to obtain lots of pedestrian images, and then involve the generated samples into  training as an extra regularization term. 
Another line of works collects the real-world data from Internet to augment the original dataset. One of the pioneering work \cite{krause2016unreasonable} is to collect large number of images via searching the keywords on the online engine, \ie, Google. After removing the noisy data, the augmented dataset facilitate the model to achieve the state-of-the-art performance on several fine-grained datasets, \eg, CUBird \cite{WahCUB_200_2011}. In a similar spirit, Zheng \etal \cite{zheng2020university} exploit noisy photos of university buildings from Google, benefiting the model learning. In contrast with these existing works, we focus on leveraging the public datasets with different data biases to learn the common knowledge given that vehicles
share the similar structure.  

\subsection{Transfer Learning}
Transfer learning is to propagate the knowledge of the source domain to the target domain \cite{pan2009survey}. On one hand, several recent works focus on the alignment between the source domain and the target domain, which intend to minimize the discrepancy of two domains. One of the pioneering works \cite{hoffman2017cycada} is to apply the cyclegan \cite{zhu2017toward} to transfer the image style to the target domain, and then train the model on the transferred data. In this way, the model could learn the similar patterns of the target data. Besides the pixel-level alignment, some works \cite{tsai2018learning,tsai2019domain,luo2019taking} focus on aligning the network activation in the middle or high layers of the neural network. The discriminator is deployed to discriminate the learned feature of source domain from that of target domain, and the main target is to minimize the feature discrepancy via adversarial learning. On the other hand, some works deploy the pseudo label learning, yielding competitive results as well \cite{zou2018unsupervised,lee2018diverse}. The main idea is to make the model more confident to the prediction, which minimizes the information entropy. The pseudo label learning usually contains two steps. The first step is to train one model from scratch on the source domain and generate the pseudo label for the unlabeled data. The second step is to fine-tune the model and make the model adapt to the target data distribution via the pseudo label. Inspired by the existing works, we propose one simple yet  effective two-stage progressive learning. We first train the model on the large-scale VehicleNet dataset and then finetune the model on the target dataset. The proposed method is also close to the traditional pre-training strategy, but the proposed method could converge quickly and yield competitive performance due to the related vehicle knowledge distilled in the model.

\section{Dataset Collection and Task Definition} \label{sec:datasetcollection}
\subsection{Dataset Analysis} \label{sec:data-analysis}
\begin{figure}[t]
\begin{center}
   \includegraphics[width=1\linewidth]{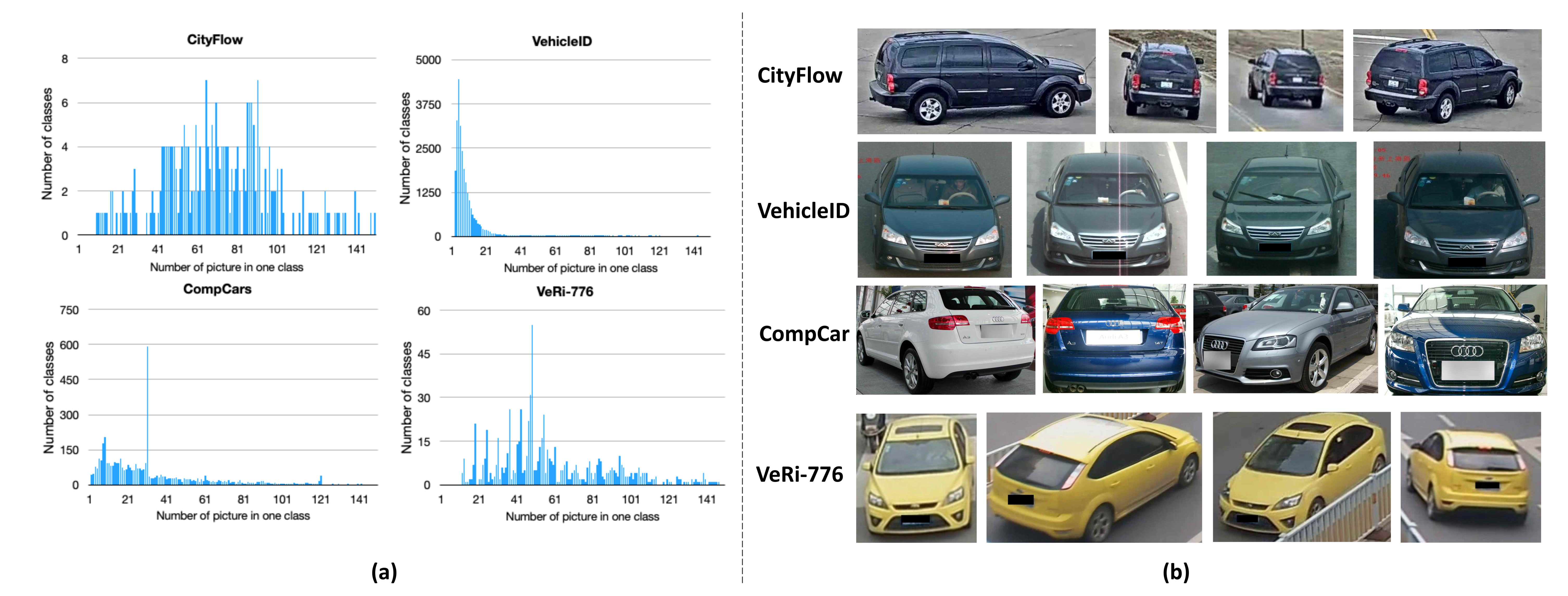}
\end{center}
\vspace{-.2in}
   \caption{(a) The image distribution per class in the vehicle re-id datasets, \eg, CityFlow \cite{tang@cityflow}, VehicleID \cite{liu2016pku} , CompCar \cite{yang2015large} and VeRi-776 \cite{liu2016deep}. We observe that the two largest datasets, \ie, VehicleID and CompCars, suffer from the limited images per class. %Note that there are only a few classes with more than 40 training images. 
   (b) Here we also provide the image samples of the four datasets. The four datasets contain different visual biases, such as  illumination conditions, collection places and viewpoints. }
\label{fig:data}
\vspace{-.1in}
\end{figure}

We involve four public datasets, \ie, CityFlow \cite{tang@cityflow}, VeRi-776 \cite{liu2016deep}, CompCar \cite{yang2015large} and VehicleID \cite{liu2016pku} into training. It results in 434,440 training images of  31,805 classes as \textbf{VehicleNet}. Note that four public datasets are collected in different places. There are no overlapping images with the validation set or the private test set. We plot the data distribution of all four datasets in Figure \ref{fig:data}. 
\textbf{CityFlow} \cite{tang@cityflow} is one of the largest vehicle re-id datasets. There are bounding boxes of 666 vehicle identities annotated. All images are collected from 40 cameras in a realistic scenario. We follow the official training/test protocol, which results in 36,935 training images of 333 classes and 19,342 testing images of other 333 classes. The training set is collected from 36 cameras, and test is collected from 23 cameras. There are 19 overlapping cameras. Official protocol does not provide a validation set. We therefore further split the training set into a validation set and a small training set. After the split, the training set contains 26,803 images of 255 classes, and the validation query set includes 463 images of the rest 78 classes. We deploy the original training set as the gallery of the validation set.  \textbf{VeRi-776} \cite{liu2016deep} contains 49,357 images of 776 vehicles from 20 cameras. The dataset is collected in the real traffic scenario, which is close to the setting of CityFlow. The author also provides the meta data, \eg, the collected time and the location. %We involve all data to train our model.
\textbf{CompCar} \cite{yang2015large} is designed for the fine-grained car recognition. It contains 136,726 images of 1,716 car models. The author provides the vehicle bounding boxes. By cropping and ignoring the invalid bounding boxes, we finally obtain 136,713 images for training. The same car model made in different years may contain the color and shape difference. We, therefore, view the same car model produced in the different years as different classes, which results in 4,701 classes. % We add all data to the  training set. 
\textbf{VehicleID} \cite{liu2016pku} consists 2211,567 images of 26,328 vehicles. The vehicle images are collected in two views, \ie, frontal and rear views. Despite the limited viewpoints, the experiment shows that VehicleID also helps the viewpoint-invariant feature learning. 
\textbf{Other Datasets} We also review other public datasets of vehicle images in Table \ref{table:datasets}. Some datasets contain limited images or views, while others lack ID annotations. %For example, PKU-VD1~\cite{yan2017exploiting} only contains the front view of cars.
Therefore, we do not use these datasets, which may potentially compromise the feature learning.

\begin{table}[!t]
\caption{Publicly available vehicle datasets. $^\dagger$: We view the vehicle model produced in different years as different classes, which leads to more classes. $^\ddagger$: The downloaded image number is slightly different with the report number in \cite{liu2016pku}.}
\vspace{-.1in}
\label{table:datasets}
\centering
\setlength{\tabcolsep}{9pt}
\begin{tabular}{l|c|c|c}
%\small
\shline
Datasets & \# Cameras & \# Images & \#IDs \\
\hline
CityFlow \cite{tang@cityflow} & 40 & 56,277 & 666 \\
VeRi-776 \cite{liu2016deep} & 20 & 49,357 & 776 \\
CompCar \cite{yang2015large} $^\dagger$ & n/a & 136,713 & 4,701 \\
VehicleID \cite{liu2016pku}  $^\ddagger$ & 2 & 221,567 & 26,328\\
PKU-VD1 \cite{yan2017exploiting} & 1 & 1,097,649 & 1,232  \\
PKU-VD2 \cite{yan2017exploiting} & 1 & 807,260 & 1,112  \\
VehicleReID \cite{Zapletal2016} & 2 & 47,123 & n/a \\
PKU-Vehicle \cite{bai2018group} & n/a & 10,000,000 & n/a \\
%Vehicle-1M \cite{guo2018learning} & 2 & 936,051 & 55,527\\
StanfordCars \cite{KrauseStarkDengFei-Fei_3DRR2013} & n/a & 16,185 & 196\\
\hline
VehicleNet & \textbf{62} & 434,440 & \textbf{31,805} \\
\shline
\end{tabular}
\end{table}

\subsection{Task Definition}
Vehicle re-identification aims to learn a projection function $F$, which maps the input image $x$ to the discriminative representation $f_i=F(x_i)$. Usually, $F$ is decided by minimizing the following optimization function on a set of training data $X = \{x_i\}_{i=1}^N$ with the annotated label $Y = \{y_i\}_{i=1}^N$:
\begin{equation}
min \sum^N_{i=1} loss(WF(x_i), y_i) + \alpha \Omega(F),
\end{equation}
where $loss(\cdot,\cdot)$ is the loss function, $W$ is the weight of the classifier, $\Omega(F)$ is the regularization term, and $\alpha$ is the weight of the regularization.

Our goal is to leverage the augmented dataset for learning robust image representation given that the vehicle shares the common structure. The challenge is to build the vehicle representation which could fit the different data distribution among multiple datasets. Given $X^d = \{x_i^d\}_{i=1}^N$ with the annotated label $Y^d = \{y_i^d\}_{i=1,d=1}^N$, the objective could be formulated as:
\begin{equation}
min \sum^D_{d=1}\sum^N_{i=1} loss(WF(x^d_i), y^d_i) + \alpha \Omega(F),
\end{equation}
where $D$ is the number of the augmented datasets. The loss demands $F$ could be applied to not only the target dataset but also other datasets, yielding the good scalability. In terms of the regularization term $\Omega(F)$, we adopt the common practise of weight decay as weight regularization, which prevents the weight value from growing too large and over-fits the dataset. 

\begin{figure}[t]
\begin{center}
   \includegraphics[width=1\linewidth]{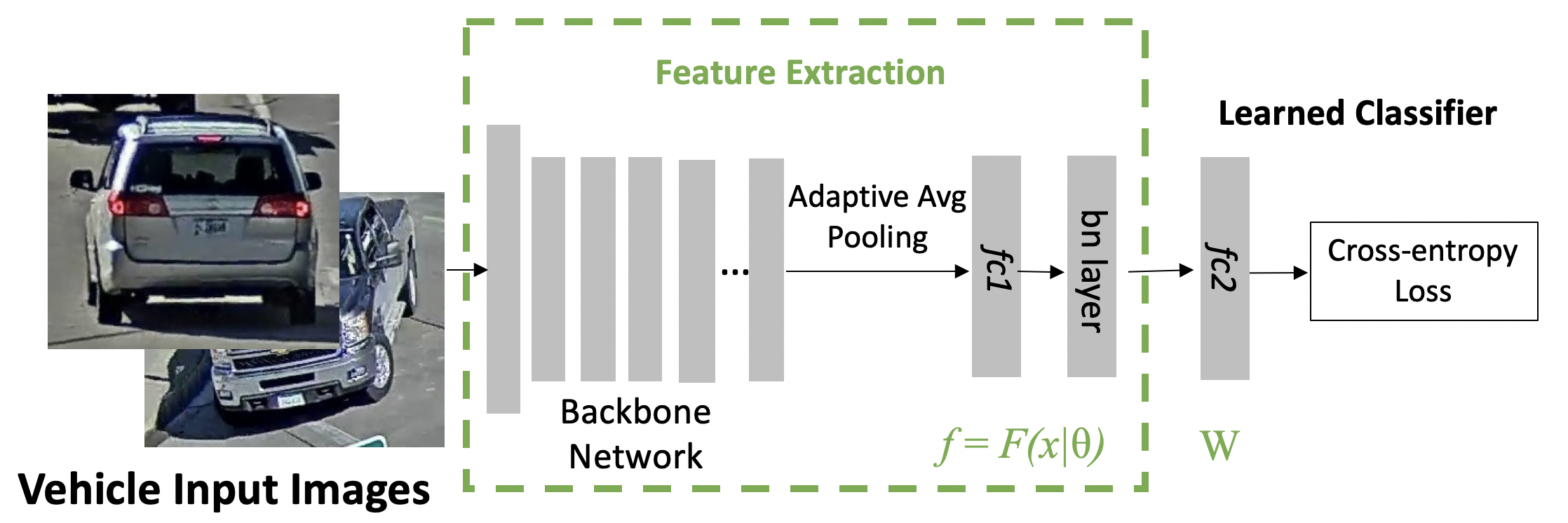}
\end{center}
\vspace{-.2in}
   \caption{Illustration of the model structure. We remove the original classifier of the ImageNet pre-trained model, add a new classifier and replace the average pooling with the adaptive average pooling layer. The adaptive average pooling is to squeeze the output to the pre-defined shape (\ie, $1\times1$). %(More details can be found in Section \ref{sec:structure}).
   }
\label{fig:structure}
\vspace{-.1in}
\end{figure}

\section{Methodology} \label{sec:method}
%We first illustrate the model structure in the Section \ref{sec:structure}. In the Section \ref{sec:two-stage}, we then introduce the proposed two-stage progressive learning method and discuss the advantage of the training strategy, followed by the description of the post-processing methods in the Section \ref{sec:post-process}.

\subsection{Model Structure} 
\label{sec:structure}
\noindent\textbf{Feature Extractor.} Following the common practise in  re-identification problems \cite{liu2016deep,zheng@joint}, we deploy the off-the-shelf Convolutional  Neural Network (CNN) model pre-trained on the ImageNet dataset \cite{russakovsky2015imagenet} as the backbone. Specifically, the proposed method is scalable and could be applied to different network backbones. We have trained and evaluated the state-of-the-art structures, including ResNet-50~\cite{he2018deep}, DenseNet-121~\cite{huang2017densely}, SE-ResNeXt101~\cite{hu2018senet} and SENet-154~\cite{hu2018senet}, in the Section \ref{sec:experiment}. %To be simplify, we deploy SE-ResNeXt101 as the example to illustrate our approach. 
The classification layer of the pre-trained backbone model is removed, which is dedicated for image recognition on ImageNet. The original average pooling layer is replaced with the adaptive average pooling layer, and the adaptive average pooling layer outputs the mean of the input feature map in terms of the height and width channels. We add one fully-connected layer `\emph{fc1}' of $512$ dimensions and one batch normalization layer to reduce the feature dimension, followed by a fully-connected layer `\emph{fc2}' to output the final classification prediction as shown in the Figure~\ref{fig:structure}. The length of the classification prediction equals to the category number of the dataset. The cross-entropy loss is to penalize the wrong vehicle category prediction.

\noindent\textbf{Feature Embedding.} Vehicle re-identification is to spot the vehicle of interest from different cameras, which demands a robust representation to various visual variants, \eg, viewpoints, illumination and resolution. Given the input image $x$, we intend to obtain the feature embedding $f = F(x|\theta)$. In this work, the CNN-based model contains the projection function $F$ and one linear classifier. Specifically, we regard the `\emph{fc2}' as the conventional linear classifier with the learnable weight $W$, and the module before the final classifier as $F$ with the learned parameter $\theta$. 
The output of the batch normalization layer as $f$ (see the green box in the Figure~\ref{fig:structure}).
When inference, we extract the feature embedding of query images and gallery images. The ranking list is generated according to the similarity with the query image. Given the query image, we deploy the cosine similarity, which could be formulated as $s(x_n, x_m) = \frac{f_n}{||f_n||_2} \times \frac{f_m}{||f_m||_2}$. The $||.||_2$ denotes $l^2$ norm of the corresponding feature embedding. The large similarity value indicates that the two images are highly relevant.

\subsection{Two-stage Progressive Learning} \label{sec:two-stage}
The proposed training strategy contains two stages. During the first stage, we train the CNN-based model on the VehicleNet dataset and learn the general representation of the vehicle images. In particular, we deploy the widely-adopted cross-entropy loss in the recognition tasks, and the model learns to identify the input vehicle images from different classes. The loss could be formulated as:
\begin{equation}
     L_{ce} = \sum^{N}_{i=1} -p_i \log (q_i) ,
      \label{eq:stage2}
\end{equation}
where $p_i$ is the one-hot vector of the ground-truth label $y_i$. The one-hot vector $p_i(c)=1$ if the index $c$ equals to $y_i$, else $p_i(c)=0$. $q_i$ is the predicted category probability of the model, and $q_i = WF(x_i|\theta)$.
Since we introduce the multi-source dataset, the cross-entropy loss could be modified to work with the multi-source data.
\begin{equation}
     L_{ce} = \sum^{D}_{d=1}\sum^{N}_{i=1}-p^d_i \log (q_i^d),
     \label{eq:stage1}
\end{equation}
where $d$ denotes the index of the public datasets in the proposed VehicleNet. Specifically, $d=1,2,3,4$ denotes the four datasets in VehicleNet, \ie, CityFlow~\cite{tang@cityflow}, VehicleID~\cite{liu2016pku} , CompCar~\cite{yang2015large} and VeRi-776~\cite{liu2016deep}, respectively. $p^d_i$ is the one-hot vector of $y^d_i$, and $q^d_i = WF(x^d_i|\theta)$. Note that we treat all the dataset equally, and demand the model with good scalability to data of different datasets in VehicleNet. 

In the first stage, we optimize the Equation~\ref{eq:stage1} on all the training data of VehicleNet to learn the shared representation for vehicle images. The Stage-I model is agnostic to the target environment, hence the training domain and the target domain are not fully aligned.
In the second stage, we take one more step to further fine-tune the model only upon the target dataset, \eg, CityFlow~\cite{tang@cityflow}, according to the Equation \ref{eq:stage2}. In this way, the model is further optimized for the target environment.
Since only one dataset is considered in the Stage-II and the number of vehicle category is decreased, in particular, the classifier is replaced with the new $fc2$ layer with $333$ classes from CityFlow.
To preserve the learned knowledge, only the classification layer of the trained model is replaced. Although the new classifier is learned from scratch, attribute to the decent initial weights in the first stage, the model could converge quickly and meets the demand for quick domain adaptation. We, therefore, could stop the training at the early epoch. To summarize, we provide the training procedure of the proposed method in Algorithm \ref{alg:RECT}. 

\begin{figure}[t]
\begin{center}
   \includegraphics[width=1\linewidth]{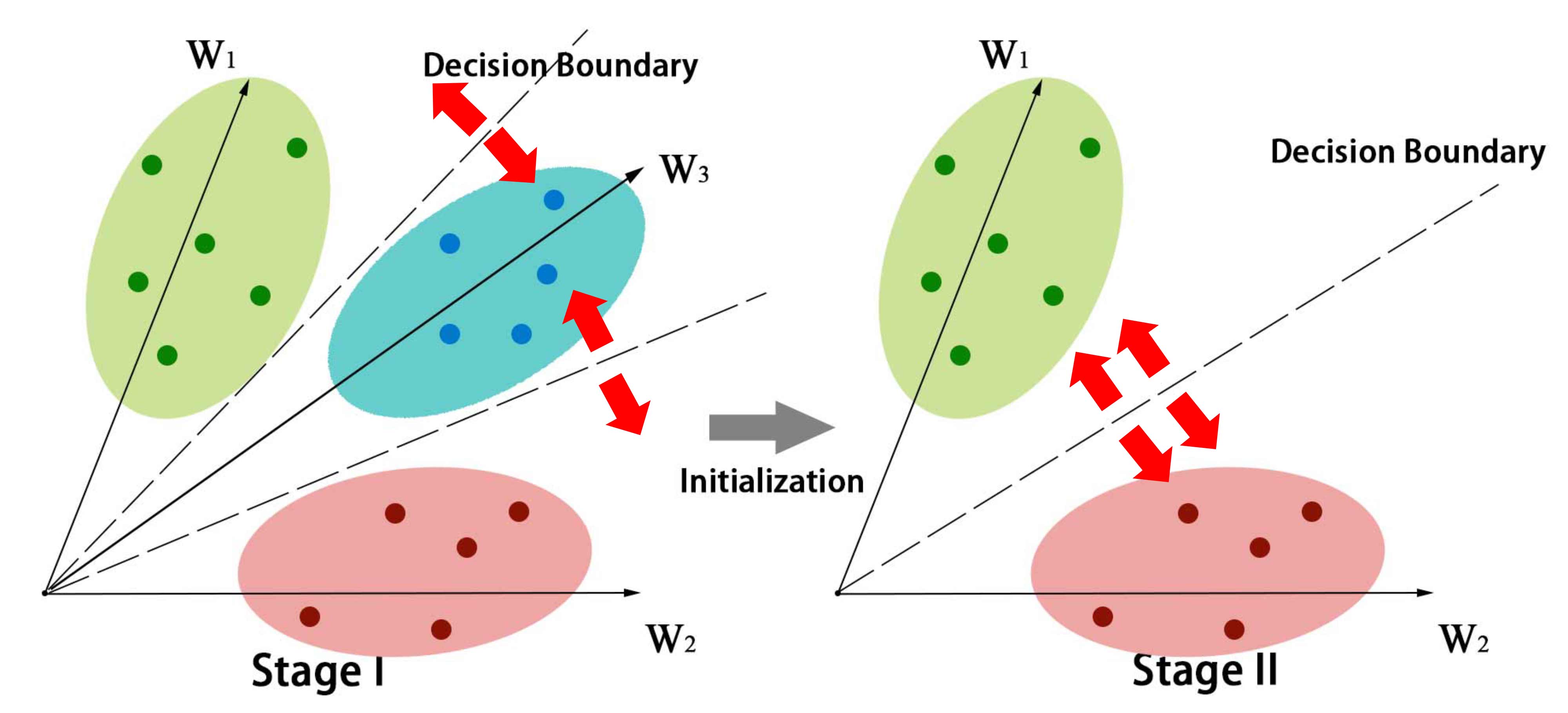}
\end{center}
\vspace{-.2in}
   \caption{Geometric Interpretation. Here we give a three-class sample to show our intuition. \zznote{$W_i$ denotes the class weight of the final linear classifier. In this example, the third class denotes one auxiliary class, which belongs to VehicleNet but the target domain. Therefore, in the Stage-II fine-tuning, we remove the auxiliary classes, including $W_3$. }
   The cross-entropy loss of Stage-I pulls the samples with the same label together (close to either the relative weight $W_1$, $W_2$ or $W_3$). In this way, the positive pair is closer than the negative pair, while the samples are far from the decision boundary. Stage I, therefore, leads to a decent weight initialization to be used in Stage II with a large margin from decision boundary, when we leave out the auxiliary class, \ie, the third class with $W_3$, from VehicleNet. }
\label{fig:advantage}
\end{figure}

\noindent\textbf{Discussion:} What are the advantages of the proposed two-stage progressive learning? First, the learned representation is more robust. In the Stage-I, we demand the model could output the discriminative representation for all of the data in the multi-source VehicleNet. The model is forced to learn the shared knowledge among the training vehicle images, which is similar to the pre-training practise in many re-ID works \cite{zheng2018pedestrian,hermans2017defense}. 
Second, the representation is also more discriminative. The first stage contains $31,805$ training classes during training. The axuiliary classes of other real vehicles could be viewed as ``virtual class'' as discussed in \cite{chen2018virtual}. Here we provide one geometric interpretation in the Figure~\ref{fig:advantage}. After the convergence of Stage I, the cross-entropy loss pulls the data with the same label together, and pushes the data from different labels away from each other on the either side of the decision boundary. In this manner, as shown in the Figure~\ref{fig:advantage}~(right), the first stage will provide better weight initialization for the subsequent fine-tuning on the target dataset. It is because the auxiliary classes expand the decision space and the data is much far from the new decision boundary, yielding discriminative features.

\begin{algorithm}[t]
\small
\caption{Training Procedure of the Proposed Method}
\label{alg:RECT}
\begin{algorithmic}[1]
\Require The multi-source VehicleNet dataset $X^d=\{x_i^d\}^D_{i=1}$; The corresponding label $Y^d=\{y_i^d\}^D_{i=1}$; 
\Require The initialized model parameter $\theta$; The first stage iteration number $T_1$ and the second stage iteration number $T_2$.
\For {$iteration = 1$ to $T_1$}
\State Stage-I: Input $x_t^j$ to $F(\cdot|\theta)$, extract the prediction of the classifier, and calculate the cross-entropy loss according to Equation \ref{eq:stage1}:
\vspace{-1.5ex}
\begin{equation}
L_{ce} = \sum^{D}_{d=1}\sum^{N}_{i=1}-p^d_i \log (q_i^d),
\vspace{-2ex}
\end{equation}
where $p^d_i$ is the one-hot vector of $y^d_i$, and $q_i^d$ is the predict probability. $q_i^d = WF(x^d_i|\theta)$, $W$ is the final fully-connected layer, which could be viewed as a linear classifer. We update the $\theta$ and $W$ during the training.
\EndFor 
\For {$iteration = 1$ to $T_2$}
\State Stage-II: We further fine-tune the trained model only on the target dataset, \eg, CityFlow. The classifier is replaced with a new one, since we have less classes. We assume that CityFlow is the first dataset ($d=1$). Thus, we could update $\theta$ upon the cross-entropy loss according to Equation \ref{eq:stage2}:
\vspace{-1.5ex}
\begin{equation}
L_{ce} = \sum^{N}_{i=1} -p^1_i \log (q^1_i).
\vspace{-2ex}
\end{equation}
where $p^1_i$ is the one-hot vector of $y^1_i$ of the CityFlow dataset, and $q_i^1$ is the predict probability. $q_i^1 = W'F(x^1_i|\theta)$. We note that $W'$ is the new fully-connected layer, which is trained from scratch and different from $W$ used in the Stage-I.
%\State We combine the prediction variance with the conventional objective to obtain the rectified objective. Update the $\theta_t$ according to Equation \ref{eq:rectified}:
%\vspace{-1.5ex}
%\begin{equation}
%L_{rect} =  \E [  exp\{-D_{kl}\} L_{ce} +  D_{kl} ]
%\end{equation}
%\vspace{-4ex}
\EndFor \\
\Return $ \theta $.
\end{algorithmic}
\end{algorithm}

\begin{figure}[t]
\begin{center}
   \includegraphics[width=1\linewidth]{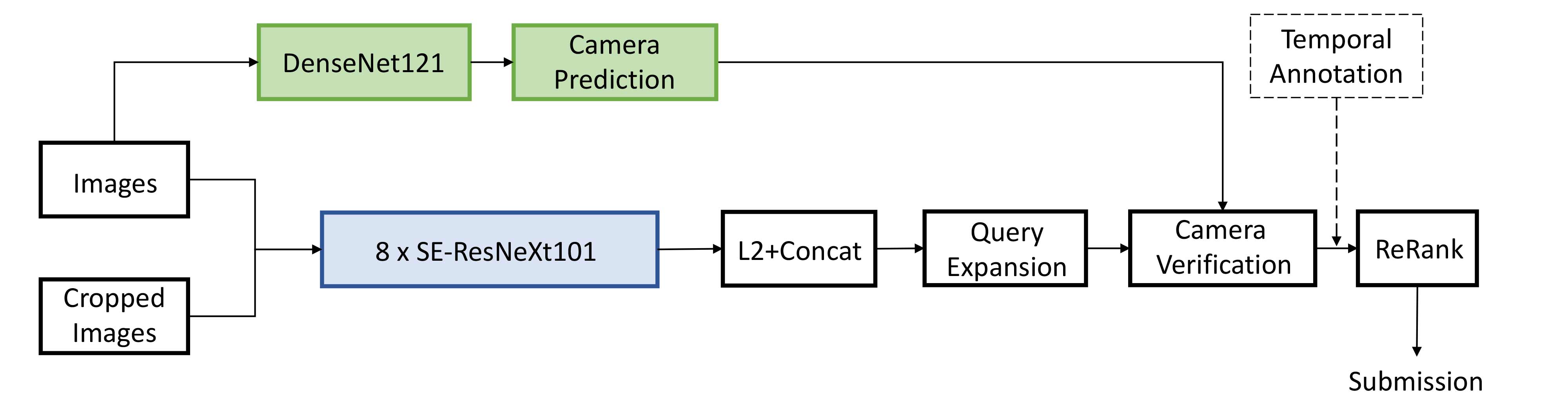}
\end{center}
\vspace{-.2in}
   \caption{The inference pipeline for AICity Challenge Competition. Given one input image and the corresponding cropped image via MaskRCNN~\cite{He_2017_ICCV}, we extract features from the trained models, \ie, $8\times$SE-ResNeXt101~\cite{hu2018senet}. We normalize and concatenate the features. Meanwhile, we extract the camera prediction from the camera-aware model, \ie, the fine-tuned DenseNet121~\cite{huang2017densely}. Then query expansion and camera verification are applied. Finally, we utilize the re-ranking technique~\cite{zhong2017re} to retrieve more positive samples. %(This is the pipeline for the submission to the private test set on AICity Challenge 2019.)
   }
\label{fig:pip}
\vspace{-.2in}
\end{figure}

\subsection{Post-processing} \label{sec:post-process}
%\textbf{Multiple Scales.} We resize the input with the factor of $[1.2, 1.1, 1.0]$. We use the mean feature of different scales.
Furthermore, we apply several post-processing techniques during the inference stage as shown in Figure \ref{fig:pip}. %For a fair comparison, we do not leverage such methods to comparing the results on the public datasets, but apply them to the AICity Challenge Competition. Next we provide some brief illustrations of the motivation as well as the mechanism of these techniques.

\noindent \textbf{Cropped Images.} We notice that the vehicle datasets usually provide a relatively loose bounding box, which may introduce the background noise. Therefore, we re-detect vehicles with the state-of-the-art MaskRCNN \cite{He_2017_ICCV}. For the final result, the vehicle representation is averaged between original images and cropped images, yielding more robust vehicle representations.

\noindent \textbf{Model Ensemble.} We adopt a straightforward late-fusion strategy, \ie, concatenating features \cite{zheng2018pedestrian}. Given the input image $x_i$, the embedding $f_i^j$ denotes the extracted feature of $x_i$ from the $j$-th trained model. The final pedestrian descriptor could be represented as:$f_i = [\frac{f_i^1}{||f_i^1||_2}, \frac{f_i^2}{||f_i^2||_2}, ... \frac{f_i^n}{||f_i^n||_2}].$ The $||\cdot||_2$ operator denotes $l^2$-norm, and $[\cdot]$ denotes feature concatenation. 

\noindent \textbf{Query Expansion \& Re-ranking.} We adopt the unsupervised clustering method, \ie, DBSCAN \cite{ester1996density} to find the most similar samples. %DBSCAN could find the areas of high density. 
The query feature is updated to the mean feature of the other queries in the same cluster. Furthermore, we adopt the re-ranking method \cite{zhong2017re} to refine the final result, which takes the high-confidence candidate images into consideration. In this work, our method does not modify the re-ranking procedure. Instead, the proposed method obtains discriminative vehicle features that distill the knowledge from ``seeing'' various cars. With better features, re-ranking is more effective.

\noindent \textbf{Camera Verification.} We utilize the camera verification to further remove some hard-negative samples. When training, we train one extra CNN model, \ie, DenseNet121~\cite{huang2017densely}, to recognize the camera from which the photo is taken. \zznote{When testing, we extract the camera-aware features from the trained model and then cluster these features by DBSCAN \cite{ester1996density}. In this way, we could obtain clustering centers.}
We applied the prior assumption that the query image and the true matches are taken in different cameras, \zznote{indicating that the query images and true matches in the gallery usually belong to different camera clustering centers}. Given a query image, we remove the images of the same camera cluster from candidate images. 

\noindent \textbf{Temporal Annotation.} 
\zznote{Temporal annotation can be easily obtained by recording the timestamp of which the target vehicle passes by. The prior assumption is that the vehicles usually appear once in the whole camera network, indicating that the two images with long time interval belong to two different vehicles. }
%One common assumption is that the cars that re-appear with long interval are different cars. 
Given the timestamp $t$ of the query image, we filter out the image in the gallery with long interval $\tau$. As a result, we only consider the candidate images with the timestamp in $[t-\tau, t+\tau]$, which also could filter out lots of the hard-negative samples.

\begin{table}
\caption{The Rank@1 (\%) and mAP (\%) accuracy with different number of training images. Here we report the results based on the validation set we splitted. $^\dagger$ Note that we split a validation set from the training set, which leads to less training data. %We apply SE-ResNeXt101 \cite{hu2018senet} as the backbone model. 
}
\vspace{-.2in}
\label{table:moredata}
\begin{center}
{
\setlength{\tabcolsep}{5pt}
\begin{tabular}{l|c|c c}
%\small
\shline
\multirow{2}{*}{Training Datasets} & \# Training & \multicolumn{2}{c}{Performance}\\
  & Images & Rank@1 (\%)& mAP (\%)\\
\hline
CityFlow \cite{tang@cityflow} $^\dagger$ & 26,803 & 73.65 & 37.65 \\
CityFlow \cite{tang@cityflow}+ VeRi-776 \cite{liu2016deep} & +49,357 & 79.48 & 43.47\\
CityFlow \cite{tang@cityflow}+ CompCar \cite{yang2015large} & +136,713 & 83.37 & 48.71\\
CityFlow \cite{tang@cityflow}+ VehicleID \cite{liu2016pku} & +221,567 & 83.37 & 47.56\\
\shline
VehicleNet & 434,440 & \textbf{88.77} & \textbf{57.35} \\
\shline
\end{tabular}}
\end{center}
\vspace{-.1in}
\end{table}

\section{Experiment} \label{sec:experiment}
%We first illustrate the implementation details in Section \ref{sec:implement} followed by the qualitative results in Section \ref{sec:qualitative}. Furthermore, we provide the futher evaluation and discussion in Section \ref{sec:further}. 

\begin{table*}
\caption{Comparison with the state-of-the-art methods in terms of Rank@1 (\%) and mAP (\%) accuracy on the VeRi-776 dataset~\cite{liu2016deep} and the VehicleID dataset~\cite{liu2016pku}. -: denotes the conventional hand-crafted features and *: denotes that the approach utilizes the self-designed network structure. The best results are in \textbf{bold}. }
\label{table:veri}
\vspace{-.2in}
\begin{center}
{
\setlength{\tabcolsep}{3pt}
\begin{tabular}{l| c| c c | c c| c c| c c}
%\small
\shline
\multirow{2}{*}{Methods} & \multirow{2}{*}{Backbones} & \multicolumn{2}{c|}{VeRi-776} & \multicolumn{2}{c|}{VehicleID~(Small)} & \multicolumn{2}{c|}{VehicleID~(Medium)} & \multicolumn{2}{c}{VehicleID~(Large)}\\
& & mAP (\%) & Rank@1 (\%) & Rank@1 (\%) & Rank@5 (\%)  & Rank@1 (\%) & Rank@5 (\%) & Rank@1 (\%) & Rank@5 (\%) \\
\hline
LOMO \cite{liao2015person} & - & 9.78 & 23.87 & 19.74 & 32.14 & 18.95 & 29.46 & 15.26 & 25.63\\
%DGD \cite{xiao2016learning} & & 17.92 & 50.70 \\
GoogLeNet \cite{yang2015large} & GoogLeNet & 17.81 & 52.12 & 47.90 & 67.43 & 43.45 & 63.53 & 38.24 & 59.51\\
FACT \cite{liu2016deep} & - & 18.73 & 51.85 & 49.53 & 67.96 & 44.63 & 64.19 &  39.91 & 60.49\\
XVGAN \cite{zhou2017cross} & * & 24.65 & 60.20 & 52.89 & 80.84 & - & - & - & - \\
SiameseVisual \cite{shen2017learning}& *&29.48 & 41.12 & - & - & - & - & - & - \\
OIFE \cite{wang2017orientation} & * &48.00 & 65.92 & - & -& -& - & 67.0 & 82.9\\
VAMI \cite{zhou2018aware}& * &50.13 & 77.03 & 63.12 & 83.25 & 52.87 & 75.12 & 47.34 & 70.29\\
NuFACT \cite{liu2017provid} & * & 53.42 & 81.56 & 48.90 & 69.51 & 43.64 & 65.34 & 38.63  & 60.72  \\
\zznote{FDA-Net} \cite{lou2019veri} &  * & 55.49 & 84.27 & - & - & 59.84 & 77.09 & 55.53 & 74.65 \\
\zznote{QD-DLF} \cite{zhu2019vehicle} & * & 61.83 & 88.50 & 72.32 & 92.48 & 70.66 & 88.90 & 68.41 & 83.37 \\
\hline
AAVER \cite{khorramshahi2019dual} & ResNet-50 & 58.52 & 88.68 & 72.47 & 93.22 & 66.85 & 89.39 & 60.23 & 84.85 \\
\zznote{PVSS} \cite{liu2019pvss} & ResNet-50 & 62.62 & 90.58 & - & - & - & - & - & - \\
\zznote{C2FRank}~\cite{guo2018learning} & GoogLeNet & - & - & 61.1 & 63.5 & 56.2 & 60.0 & 51.4 & 53.0 \\
VANet \cite{chu2019vehicle} & GoogLeNet & 66.34 & 89.78 & 83.26 & 95.97 & 81.11 & \textbf{94.71} & 77.21 & \textbf{92.92}\\
PAMTRI \cite{tang2019pamtri} & DenseNet-121 & 71.88 & 92.86 & - & - & - & - & - & - \\
SAN \cite{qian2019stripe} & ResNet-50 & 72.5 & 93.3 & 79.7 & 94.3 & 78.4 & 91.3 & 75.6 & 88.3\\ 
Part \cite{he2019part} & ResNet-50 & 74.3 & 94.3 & 78.4 & 92.3 & 75.0 & 88.3 & 74.2 & 86.4 \\
\hline 
Ours (Stage-I) & ResNet-50 & 80.91 & 95.95 &83.26 & 96.77 & 81.13 & 93.68 & 79.06 & 91.84 \\
Ours (Stage-II) & ResNet-50 & \textbf{83.41} & \textbf{96.78} &  \textbf{83.64} & \textbf{96.86} & \textbf{81.35} & 93.61 & \textbf{79.46} & 92.04 \\
\shline
\end{tabular}}
\end{center}
\vspace{-.2in}
\end{table*}

\subsection{Implementation Details} \label{sec:implement}
For two widely-adopted public datasets, \ie, VeRi-776 and VehicleID, we follow the setting in \cite{qian2019stripe,he2019part} to conduct a fair comparison. We adopt ResNet-50 \cite{he2016deep} as the backbone network and input images are resized to $256 \times 256$. We apply SGD optimizer with momentum of 0.9 and mini-batch size of $36$. The weight decay is set to 0.0001 following the setting in \cite{he2016deep}. The initial learning rate is set to $0.02$ and is divided by a factor $10$ at the $40$-th epoch of the first stage and the $8$-th epoch in the second stage. The total epochs of the first stage is $60$ epochs, while the second-stage fine-tuning is trained with $12$ epochs. \textbf{When inference, we only apply the mean feature of the image flipped horizontally, without using other post-processing approaches for two academic datasets.} 

For the competition dataset, \ie, CityFlow~\cite{tang@cityflow}, we adopt one sophisticated model, \ie, SE-ResNeXt101 \cite{hu2018senet} as the backbone to conduct the ablation study and report the performance. The vehicle images are resized to $384 \times 384$. Similarly, the first stage is trained with $60$ epochs, and the second stage contains $12$ epochs. When conducting inference on the validation set, we only apply the mean feature of the image flipped horizontally, without using other post-processing approaches. In contrast, to achieve the best results on the private test set of CityFlow, we apply all the post-processing methods mentioned in Section \ref{sec:post-process}.
%To verify the effectiveness of the proposed dataset and the approach, we conduct the ablation study and report the results of the validation set in Section \ref{sec:further}.

%\noindent\textbf{Evaluation Metric.} Following previous works \cite{tang@cityflow,tang2019pamtri}, we adopt two widely-used evaluation metrics, \ie, Rank@K and mAP. Rank@K is the probability that the true-match image appears in the top-K of the ranking list. Given a ranking list, the average precision (AP) calculates the space under the recall-precision curve, while mAP is the mean of the average precision of all queries. 

%\noindent\textbf{Reproducibility.} The code is based on Pytorch \cite{paszke2017automatic}. We will release our code for reproducibility upon the publish.

\subsection{Qualitative Results} \label{sec:qualitative}
\noindent \textbf{Effect of VehicleNet.}
To verify the effectiveness of the public vehicle data towards the model performance, we involve different vehicle datasets into training and report the results, respectively (see Table~\ref{table:moredata}). There are two primary points as follows: First, the model performance has been improved by involving the training data of one certain datasets, either VeRi-776, CompCar or VehicleID. For instance, the model trained on CityFlow + CompCar has achieved $83.37\%$ Rank@1 and $48.71\%$ mAP, which surpasses the baseline of $73.65\%$ Rank@1 and $37.65\%$ mAP. It shows that more training data from other public datasets indeed helps the model learning the robust representation of vehicle images. Second, we utilize the proposed large-scale VehicleNet to train the model, which contains all the training data of four public datasets. We notice that there are $+15.12\%$ Rank@1 improvement from $73.65\%$ Rank@1 to $88.77\%$ Rank@1, and $+19.70\%$ mAP increment from $37.65\%$ mAP to $57.35\%$ mAP. It shows that the proposed VehicleNet has successfully ``borrowed'' the strength from multiple datasets and help the model learning robust and discriminative features.

\begin{table}
\caption{Competition results of AICity Vehicle Re-id Challenge on the private test set. Our results are in \textbf{bold}.}
\vspace{-.2in}
\label{table:team}
\begin{center}
{
\setlength{\tabcolsep}{5pt}
\begin{tabular}{l | c| c}
%\small
\shline
Team Name & Temporal Annotation & mAP(\%) \\
\hline
Baidu\_ZeroOne~\cite{tan2019multi} & \checkmark &85.54  \\
UWIPL~\cite{huang2019multi}  & \checkmark & 79.17  \\
ANU~\cite{lv2019vehicle}  & \checkmark & 75.89  \\
\hline
\textbf{Ours} & $\times$ & \textbf{75.60} \\
\textbf{Ours}& \checkmark & \textbf{86.07} \\
%5 & Traffic Brain~\cite{he2019multi} & \checkmark & 73.02  \\
%6 & Desire & - & 67.93 \\
%7 & XINGZHI & - & 60.91  \\
%8 & UWD\_RC & - & 60.78 \\
%9 & MVM & - & 58.62  \\
%10 & flyZJ & & 58.27  \\
%\hline
% & Baseline~\cite{tang@cityflow} & $\times$ & 32.0 \\
\shline
\end{tabular}}
\end{center}
\vspace{-.2in}
\end{table}

\noindent \textbf{Comparison with the State-of-the-art.} We mainly compare the performance with other methods on the test sets of two public vehicle re-id datasets, \ie, VeRi-776~\cite{liu2016deep} and VehicleID~\cite{liu2016pku} as well as AICity Challenge ~\cite{tang2019pamtri} private test set.  The comparison results with other competitive methods are as follows: \textbf{VeRi-776 \& VehicleID.} There are two lines of competitive methods. One line of works deploy the hand-crafted features \cite{liao2015person,liu2016deep} or utilize the self-designed network \cite{zhou2018aware,wang2017orientation,liu2017provid}. In contrast, another line of works leverages the model pre-trained on ImageNet, yielding the superior performance \cite{khorramshahi2019dual,chu2019vehicle,tang2019pamtri,he2019part}. As shown in Table \ref{table:veri}, we first evaluate the proposed approach on the VeRi-776 dataset \cite{liu2016deep}. We leave out the VeRi-776 test set from the VehicleNet to fairly compare the performance, and we deploy the ResNet-50~\cite{he2016deep} as backbone network, which is used by most compared methods. The proposed method has achieved $83.41\%$ mAP and $96.78\%$ Rank@1 accuracy, which is superior to the second best method, \ie, Part-based model~\cite{he2019part} ($74.3\%$ mAP and $94.3\%$ Rank@1) by a large margin. %Furthermore, if we fine-tune the model in the Stage-II, the performance is further improved to $81.28\%$ mAP and $96.19\%$ Rank@1 accuracy. 
Meanwhile, we observe a similar result on the VehicleID dataset~\cite{liu2016pku} in all three settings (Small /Medium /Large). Small, Medium and Large setting denotes different gallery sizes of 800, 1600 and 2400, respectively. The proposed method also arrives competitive results, \eg, $83.64\%$ Rank@1 of the small gallery setting, $81.35\%$ Rank@1 of the medium gallery setting, and $79.46\%$ Rank@1 of the large gallery setting. \zznote{One competitive method, VANet~\cite{chu2019vehicle}, has achieved comparable results on VehicleID, but is inferior to the proposed method on VeRi-776. It is because VANet introduces one extra viewpoint module, which could discriminate different viewpoints, i.e., front view and rear view. Since the VehicleID dataset only contains two views, VANet works well. In contrast, on another benchmark VeRi-776, containing 20 cameras, the proposed method is more scalable than VANet in terms of the multi-camera scenario.} 
\textbf{AICity Challenge.} For AICity Challenge Competition (on the private test set of CityFlow \cite{tang@cityflow}), we adopt a slightly different training strategy, using the large input size as well as the model ensemble. The images are resized to $384 \times 384$. We adopt the mini-batch SGD with the weight decay of 5e-4 and a momentum of $0.9$. In the first stage, we decay the learning rate of $0.1$ at the $40$-th and $55$-th epoch. We trained $32$ models with different batchsizes and different learning rates. In the second stage, we fine-tune the models on the original dataset. We decay the learning rate of 0.1 at the $8$-th epoch and stop training at the $12$-th epoch. Finally, we select $8$ best models on the validation set to extract the feature. When testing, we adopt the horizontal flipping and scale jittering, which resizes the image with the scale factors $[1, 0.9, 0.8]$ to extract features. As a result, we arrive at 75.60\% mAP on the private testing set. Without extra temporal annotations, our method has already achieved competitive results (see Table  \ref{table:team}). With the help of extra annotation of temporal and spatial information, we have achieved $86.07\%$ mAP, which surpasses the champion of the AICity Vehicle Re-id Challenge 2019.

\begin{table}
\caption{The Rank@1(\%) and mAP (\%) accuracy with different stages on the CityFlow private test set. %We report the results on the private test set rather than validation set, since we involve all training images into fine-tuning. Post-processing methods are leveraged on the private test set.
}
\label{table:stage}
\vspace{-.2in}
\begin{center}
{
\setlength{\tabcolsep}{15pt}
\begin{tabular}{l | c c}
%\small
\shline
\multirow{2}{*}{} & \multicolumn{2}{c}{Private Test Set}\\
   & Rank@1(\%) & mAP(\%) \\
\hline
Stage I & 82.70 &  68.21 \\
Stage II & 87.45 &  75.60 \\
\shline
\end{tabular}}
\end{center}
\vspace{-.2in}
\end{table}

\begin{table}
\caption{Effect of different post-processing techniques on the CityFlow validation set.}
\label{table:post}
\vspace{-.2in}
\begin{center}
{
\setlength{\tabcolsep}{3pt}
\begin{tabular}{l|c c c c c c }
%\small
\shline
Method & \multicolumn{6}{c}{Performance} \\
\hline
with Cropped Image? & & $\checkmark$ & $\checkmark$ & $\checkmark$ & $\checkmark$ & $\checkmark$ \\
Model Ensemble? & & & $\checkmark$ & $\checkmark$ & $\checkmark$ &$\checkmark$ \\
Query Expansion? & & & & $\checkmark$ & $\checkmark$ & $\checkmark$ \\
Camera Verification? & & & & & $\checkmark$ & $\checkmark$  \\
Re-ranking?  & & & & & & $\checkmark$ \\
\hline
mAP (\%) & 57.35 & 57.68 & 61.29 & 63.97 & 65.97 & 74.52 \\
\shline
\end{tabular}}
\end{center}
\vspace{-.1in}
\end{table}

\begin{table}
\caption{The Rank@1 (\%) and mAP (\%) accuracy with different backbones on the CityFlow validation set. The best results are in \textbf{bold}. }
\label{table:backbones}
\vspace{-.2in}
\begin{center}
{
\setlength{\tabcolsep}{8pt}
\begin{tabular}{l| c| c c}
%\small
\shline
\multirow{2}{*}{Backbones} & ImageNet & \multicolumn{2}{c}{Performance}\\
 & Top5(\%) & Rank@1 (\%) & mAP (\%) \\
\hline
ResNet-50 \cite{he2016deep} & 92.98 &77.97 & 43.65 \\
DenseNet-121 \cite{huang2017densely}& 92.14 &83.15 & 47.17 \\
SE-ResNeXt101 \cite{hu2018senet} & 95.04 &\textbf{83.37} & \textbf{48.71} \\
SENet-154 \cite{hu2018senet}  & 95.53 & 81.43 & 45.14 \\
%NASNet \cite{zoph2018learning}  & &  \\
\shline
\end{tabular}}
\end{center}
\vspace{-.2in}
\end{table}

\begin{table}
\caption{The Rank@1(\%) and mAP (\%) accuracy on the CityFlow validation set with two different sampling methods. Here we use the ResNet-50 backbone.}
\label{table:sample}
\vspace{-.2in}
\begin{center}
{
\setlength{\tabcolsep}{10pt}
\begin{tabular}{l|c c}
%\small
\shline
\zznote{\multirow{2}{*}{Sampling Policy}} & \multicolumn{2}{c}{Performance}\\
  & Rank@1(\%) & mAP(\%) \\
\hline
Naive Sampling & 77.97  & 43.65 \\
Balanced Sampling  & 76.03 & 40.09 \\
\shline
\end{tabular}}
\end{center}
\vspace{-.1in}
\end{table}

\begin{figure}[t]
\begin{center}
   \includegraphics[width=1\linewidth]{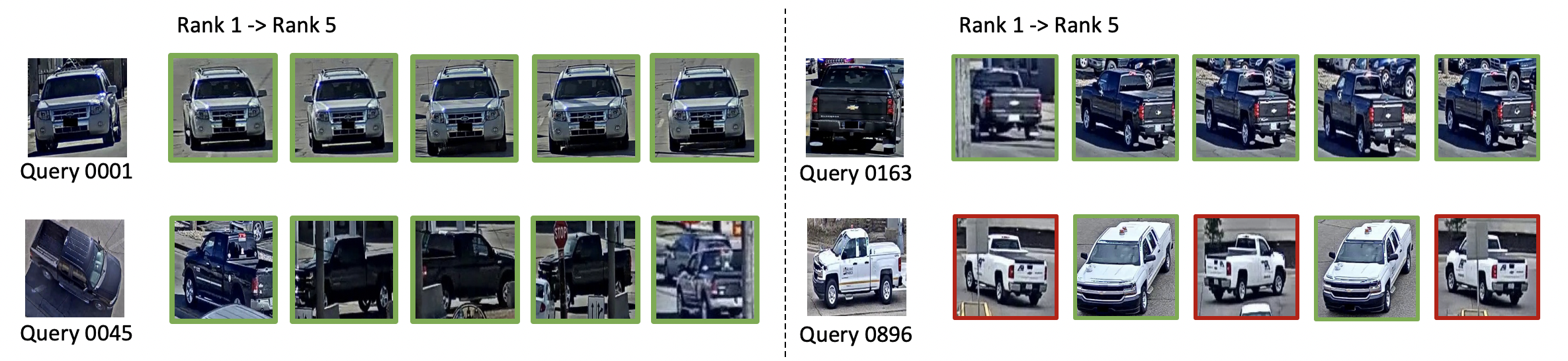}
\end{center}
\vspace{-.2in}
   \caption{Qualitative image search results using the vehicle query images from the CityFlow dataset. We select the four query images from different viewpoints. %, \ie, the front view, the overhead view, the rear view and the side view. 
   The results are sorted from left to right according to the similarity score. The true-matches are in \textcolor{ForestGreen}{green}, when the false-matches are in \textcolor{red}{red}. }
\label{fig:show}
\end{figure}

\begin{figure}[t]
\begin{center}
   \includegraphics[width=1\linewidth]{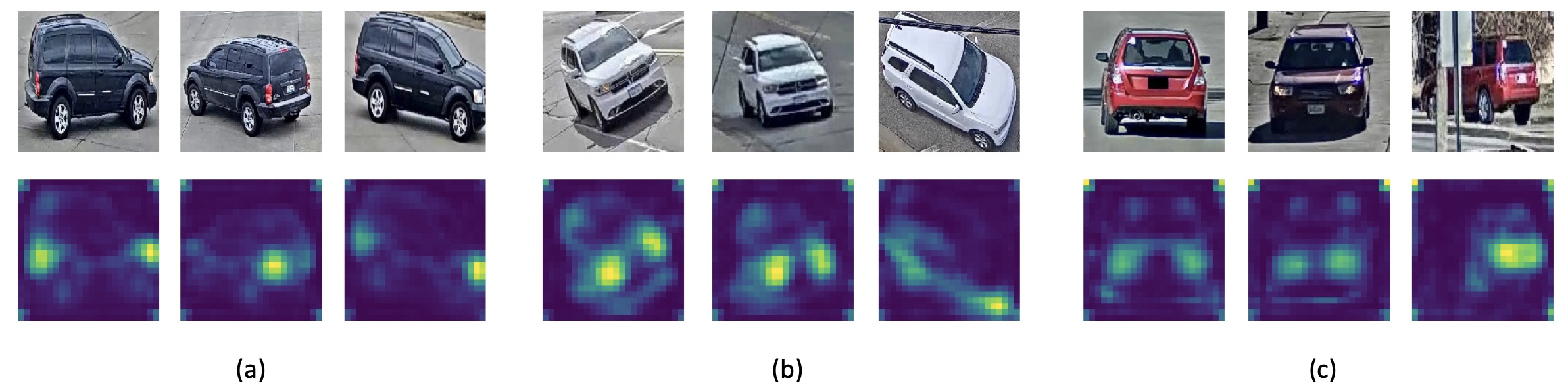}
\end{center}
\vspace{-.2in}
   \caption{Visualization of the activation heatmap in the learned model on VehicleNet. The vehicle images in every subfigure (a)-(c) are from the same vehicle ID. Noted that there do exist strong response values at the regions containing discriminative details, such as headlights and tire types. }
\label{fig:heatmap}
\vspace{-.1in}
\end{figure}

\subsection{Further Evaluations and Discussion} \label{sec:further}
\noindent \textbf{Effect of Two-stage Progressive Learning.} We compare the final results of the Stage I and the Stage II on the private test set of CityFlow (see Table \ref{table:stage}). We do not  evaluate the performance on the validation set we splitted, since we utilize all training images into fine-tuning. The model of Stage II has arrived $87.45 \%$ Rank@1 and $75.60\%$ mAP accuracy, which has significantly surpassed the one of Stage I $+7.39\%$ mAP and $+4.75\%$ Rank@1. It verifies the effectiveness of the two-stage learning. In the Stage I, the target training set, \ie, CityFlow, only occupy $6\%$ of VehicleNet. The learned model, therefore, is sub-optimal for the target environment. To further optimize the model for CityFlow, the second stage fine-tuning helps to minor the gap between VehicleNet and the target set, yielding better performance. %Besides, the second stage usually converges quickly, which also could enable the fast domain adaptation.
We also observe similar results on the other two datasets, \ie, VeRi-776 and VehicleID. As shown in the last two row of Table \ref{table:veri}, the Stage-II fine-tuning could further boost the performance. For instance, the proposed method has achieved $+2.50\%$ mAP and $+0.83\%$ Rank@1 improvement on VeRi-776. 
We compare the two-stage learning strategy with the domain adaption policy, which is usually based on style transferring. Specifically, we apply the prevailing CycleGAN~\cite{zhu2017toward} to change the style of data in VehicleNet to VeRi-776. We observe that CycleGAN could successfully change the vehicle style. However, CycleGAN introduces some unrealistic noise. As shown in Table~\ref{table:complementary}, the style transferring method is inferior to the proposed two-stage learning strategy. We speculate that it is due to the generation noise by CycleGAN. Besides, training CycleGAN costs extra time, which may be not ideal for the fast domain adaptation.

\begin{table}
\caption{\zznote{Comparison with other complementary methods on VeRi-776.}}
\vspace{-.2in}
\label{table:complementary}
\begin{center}
{
\setlength{\tabcolsep}{15pt}
\begin{tabular}{l | c c}
%\small
\shline
 Method  & Rank@1(\%) & mAP(\%) \\
\hline 
\zznote{$w$ CycleGAN data} & 92.91 & 75.23 \\ 
\hline
Stage I & 95.95 &  80.91 \\
Stage II & 96.78 &  83.41 \\
\zznote{Stage II + PCB~\cite{sun2017beyond}} & 97.26 &  83.54 \\
\shline
\end{tabular}}
\end{center}
\vspace{-.2in}
\end{table}

\noindent \textbf{Effect of Part-based Method Fusion.} The proposed method has the potential to fuse with other competitive methods. We select the second best method~\cite{qian2019stripe} on VeRi-776 to verify the potential of the proposed method. \cite{qian2019stripe} utilizes one similar policy as PCB ~\cite{sun2017beyond} to split the feature map horizontally into 4 parts. As shown in Table~\ref{table:complementary}, ours + PCB can take one step further, yielding 97.26\% Rank@1 and 83.54\% mAP.

\noindent \textbf{Effect of Post-processing.} Here we provide the ablation study of post-processing techniques on the validation set of CityFlow (see Table \ref{table:post}). When applying the augmentation with cropped images, model ensemble, query expansion, camera verification and re-ranking, the performance gradually increases, which verifies the effectiveness of post-processing methods. %We also apply the similar policy to the final result on the private test set of AICity Challenge.  

\noindent \textbf{Effect of Different Backbones.} We observe that different backbones may lead to different results. As shown in Table \ref{table:backbones}, SE-ResNeXt101 \cite{hu2018senet} arrives the best performance with $83.37$ Rank@1 and $48.71\%$ mAP on the validation set of the CityFlow dataset. We speculate that it is tricky to optimize some large-scale neural networks due to the problem of gradient vanishing. For instance, we do not achieve a better result ($45.14\%$ mAP) with SENet-154 \cite{hu2018senet}, which preforms better than SE-ResNeXt101~\cite{hu2018senet} on ImageNet \cite{deng2009imagenet}. We hope this observation could help the further study of the model backbone selection in terms of the re-identification task.

\noindent \textbf{Effect of Sampling Policy.} 
Since we introduce more training data in the first stage, the data sampling policy has a large impact on the final result. We compare two sampling policies. The naive method is to sample every image once in every epoch. Another method is called balanced sampling policy. The balanced sampling is to sample the images of different class with equal possibility. As shown in Table \ref{table:sample}, the balanced sampling harms the result. We speculate that the long-tailed data distribution (see Figure \ref{fig:data}) makes the balanced sampling have more chance to select the same image in the classes with fewer images. Thus the model is prone to over-fit the class with limited samples, which compromise the final performance. Therefore, we adopt the naive sampling policy. 

%\noindent \textbf{More Data Matters.} As shown in Table \ref{table:moredata} , involving more training data consistently improves the result. The model sees more vehicle images taken by different cameras, and learn the viewpoint-invariant features. 

\noindent \textbf{Visualization of Vehicle Re-id Results.} As shown in Figure \ref{fig:show}, we provide the qualitative image search results on CityFlow. We select the four query images from different viewpoints, \ie, the front view, the overhead view, the rear view and the side view. The proposed method has successfully retrieved the relevant results in the top-5 of the ranking list. %Since there are multiple ground-truth images from the same video sequence, the top-5 ranking results on the CityFlow is prone to be the images from the same video sequence.  

\noindent \textbf{Visualization of Learned Heatmap.} Following \cite{zheng2016discriminatively,bai2018group}, we utilize the network activation before the pooling layer to visualize the attention of the learned model. \zznote{For instance, given one middle-level feature of $14\times14\times2048$, we aggregate the activation of all channels via summation, resulting one feature of $14\times14$. Then we normalize the feature to [0,1], and map the value to the corresponding heatmap color. The generation code is avaiable at \footnote{\tiny\url{https://github.com/layumi/Person_reID_baseline_pytorch/blob/dev/visual_heatmap.py  }}. }As shown in Figure \ref{fig:heatmap}, the trained model has strong response values at the regions containing discriminative details, such as headlights and tire types. In particular, despite different viewpoints, the model could focus on the salient areas, yielding the viewpoint-invariant feature.

\begin{figure}[t]
\begin{center}
   \includegraphics[width=0.95\linewidth]{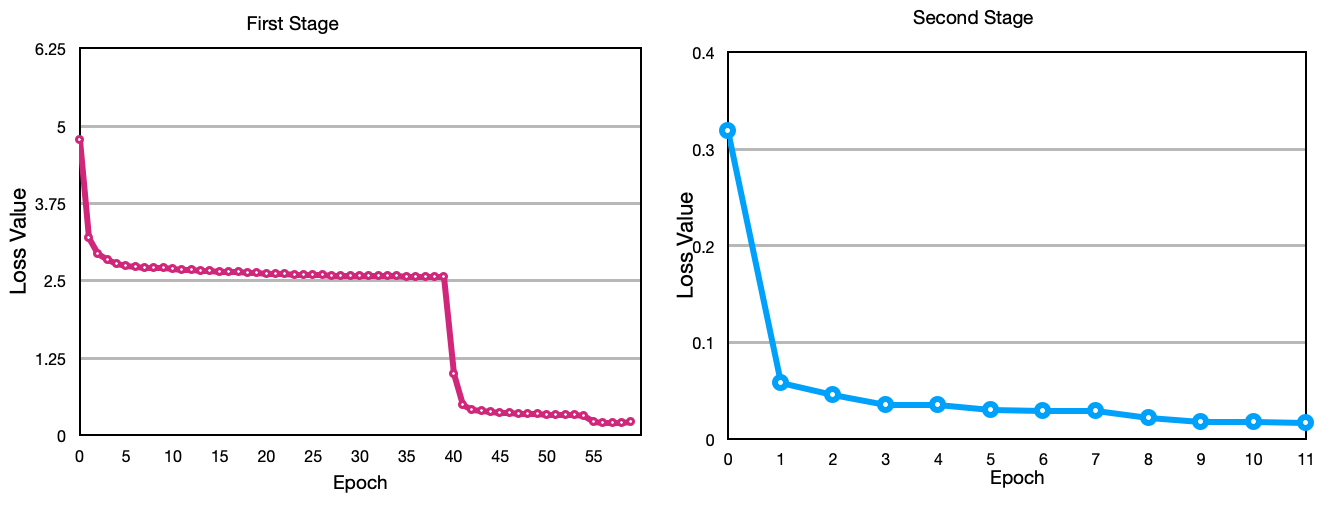}
\end{center}
\vspace{-.2in}
   \caption{The training losses of the two stages. Due to the large-scale data and classes, the first stage (left) takes more epochs to converge. Attribute to the trained weight of the first stage, the second stage (right) converge early.}
\label{fig:loss}
\vspace{-.1in}
\end{figure}

\noindent \textbf{Model Convergence.} As shown in Figure \ref{fig:loss} (left), despite a large number of training classes, \ie, $31,805$ categories in VehicleNet, the model could converge within $60$ epochs. As discussed, the first stage provides a decent weight initialization for fine-tuning in the second stage. Therefore, Stage-II training converges quickly within 12 epochs (see Figure~\ref{fig:loss} (right)). 

\noindent \textbf{Time Cost.} The Stage-I training costs about 30 hours on the whole VehicleNet with 3 $\times$ Nvidia 2080TI. The Stage-II training costs about 1.5 hours for fine-tuning.

\section{Conclusion} \label{sec:conclusion}
In this paper, we intend to address two challenges in the context of vehicle re-identification, \ie, the lack of training data, and how to harness multiple public datasets. To address the data limitation, we build a large-scale dataset called VehicleNet. % with free vehicle training images from public datasets. 
To learn the robust feature, we propose a simple yet  effective approach, called two-stage progressive learning. %, and discuss advantages of the learning strategy. 
We %evaluate the method on the private test set of CityFlow~\cite{tang@cityflow} and 
achieve $86.07\%$ mAP accuracy in AICity19 Challenge %The proposed method has surpassed the champion of the challenge, yielding $86.07\%$ mAP accuracy. Besides, the proposed method also achieves
and competitive performance on two other public datasets, \ie, VeRi-776 and VehicleID.
%In this paper, we show that more training data matters, and could contribute to the learning of robust visual representation. However, the data collection is still challenging.
In the future, we will try two data collection methods to further improve the work. 1). One method is to collect data from the search engine, \ie, Google, to enlarge the dataset. The existing works \cite{krause2016unreasonable,zheng2020university} show that a few noise annotations usually do not compromise the model training. %, while most true-annotated data plays the important role to improve the performance. 
2). The other way is to generate the synthetic data by either GAN~\cite{goodfellow2014generative} or 3D-models~\cite{yao2019simulating}, to further explore the robust representation learning. %After high-quality data collection, 
Besides, we will explore weakly supervised learning approaches~\cite{zhang2019leveraging,zhang2018spftn,meng2019weakly} to fully take advantage of unlabeled data.

%\appendices

%\section{}

%\section*{Acknowledgment}


\ifCLASSOPTIONcaptionsoff
  \newpage
\fi


\bibliographystyle{IEEEtran}
\bibliography{IEEEabrv,mybib}

\begin{thebibliography}{10}
\providecommand{\url}[1]{#1}
\csname url@samestyle\endcsname
\providecommand{\newblock}{\relax}
\providecommand{\bibinfo}[2]{#2}
\providecommand{\BIBentrySTDinterwordspacing}{\spaceskip=0pt\relax}
\providecommand{\BIBentryALTinterwordstretchfactor}{4}
\providecommand{\BIBentryALTinterwordspacing}{\spaceskip=\fontdimen2\font plus
\BIBentryALTinterwordstretchfactor\fontdimen3\font minus
  \fontdimen4\font\relax}
\providecommand{\BIBforeignlanguage}[2]{{%
\expandafter\ifx\csname l@#1\endcsname\relax
\typeout{** WARNING: IEEEtran.bst: No hyphenation pattern has been}%
\typeout{** loaded for the language `#1'. Using the pattern for}%
\typeout{** the default language instead.}%
\else
\language=\csname l@#1\endcsname
\fi
#2}}
\providecommand{\BIBdecl}{\relax}
\BIBdecl

\bibitem{zheng2019joint}
Z.~Zheng, X.~Yang, Z.~Yu, L.~Zheng, Y.~Yang, and J.~Kautz, ``Joint
  discriminative and generative learning for person re-identification,''
  \emph{CVPR}, 2019.

\bibitem{tang@cityflow}
T.~Zheng, N.~Milind, L.~Ming-Yu, Y.~Xiaodong, B.~Stan, W.~Shuo, K.~Ratnesh,
  A.~David, and H.~Jenq-Neng, ``Cityflow: A city-scale benchmark for
  multi-target multi-camera vehicle tracking and re-identification,'' in
  \emph{CVPR}, 2019.

\bibitem{sun2017beyond}
Y.~Sun, L.~Zheng, Y.~Yang, Q.~Tian, and S.~Wang, ``Beyond part models: Person
  retrieval with refined part pooling,'' in \emph{ECCV}, 2018.

\bibitem{sun2017svdnet}
Y.~Sun, L.~Zheng, W.~Deng, and S.~Wang, ``{SVDNet} for pedestrian retrieval,''
  in \emph{ICCV}, 2017.

\bibitem{zheng2018pedestrian}
Z.~Zheng, L.~Zheng, and Y.~Yang, ``Pedestrian alignment network for large-scale
  person re-identification,'' \emph{IEEE Transactions on Circuits and Systems
  for Video Technology}, 2018.

\bibitem{liu2016deep}
X.~Liu, W.~Liu, T.~Mei, and H.~Ma, ``A deep learning-based approach to
  progressive vehicle re-identification for urban surveillance,'' in
  \emph{ECCV}, 2016.

\bibitem{zhou2018aware}
Y.~Zhou and L.~Shao, ``Aware attentive multi-view inference for vehicle
  re-identification,'' in \emph{CVPR}, 2018.

\bibitem{wang2017orientation}
Z.~Wang, L.~Tang, X.~Liu, Z.~Yao, S.~Yi, J.~Shao, J.~Yan, S.~Wang, H.~Li, and
  X.~Wang, ``Orientation invariant feature embedding and spatial temporal
  regularization for vehicle re-identification,'' in \emph{ICCV}, 2017.

\bibitem{deng2009imagenet}
J.~Deng, W.~Dong, R.~Socher, L.-J. Li, K.~Li, and L.~Fei-Fei, ``Imagenet: A
  large-scale hierarchical image database,'' in \emph{CVPR}, 2009.

\bibitem{yang2015large}
L.~Yang, P.~Luo, C.~Change~Loy, and X.~Tang, ``A large-scale car dataset for
  fine-grained categorization and verification,'' in \emph{CVPR}, 2015.

\bibitem{pan2009survey}
S.~J. Pan and Q.~Yang, ``A survey on transfer learning,'' \emph{TKDE}, vol.~22,
  no.~10, pp. 1345--1359, 2009.

\bibitem{krause2016unreasonable}
J.~Krause, B.~Sapp, A.~Howard, H.~Zhou, A.~Toshev, T.~Duerig, J.~Philbin, and
  L.~Fei-Fei, ``The unreasonable effectiveness of noisy data for fine-grained
  recognition,'' in \emph{ECCV}, 2016.

\bibitem{mahajan2018exploring}
D.~Mahajan, R.~Girshick, V.~Ramanathan, K.~He, M.~Paluri, Y.~Li, A.~Bharambe,
  and L.~van~der Maaten, ``Exploring the limits of weakly supervised
  pretraining,'' in \emph{ECCV}, 2018.

\bibitem{liu2016pku}
H.~Liu, Y.~Tian, Y.~Wang, L.~Pang, and T.~Huang, ``Deep relative distance
  learning: Tell the difference between similar vehicles,'' in \emph{CVPR},
  2016.

\bibitem{yan2017exploiting}
K.~Yan, Y.~Tian, Y.~Wang, W.~Zeng, and T.~Huang, ``Exploiting multi-grain
  ranking constraints for precisely searching visually-similar vehicles,'' in
  \emph{ICCV}, 2017.

\bibitem{zapletal2016vehicle}
D.~Zapletal and A.~Herout, ``Vehicle re-identification for automatic video
  traffic surveillance,'' in \emph{CVPR Workshops}, 2016.

\bibitem{liu2016large}
W.~Liu, Y.~Wen, Z.~Yu, and M.~Yang, ``Large-margin softmax loss for
  convolutional neural networks,'' in \emph{ICML}, 2016.

\bibitem{qian2017multi}
X.~Qian, Y.~Fu, Y.-G. Jiang, T.~Xiang, and X.~Xue, ``Multi-scale deep learning
  architectures for person re-identification,'' \emph{CVPR}, 2017.

\bibitem{guo2018learning}
H.~Guo, C.~Zhao, Z.~Liu, J.~Wang, and H.~Lu, ``Learning coarse-to-fine
  structured feature embedding for vehicle re-identification,'' in \emph{AAAI},
  2018.

\bibitem{shen2017learning}
Y.~Shen, T.~Xiao, H.~Li, S.~Yi, and X.~Wang, ``Learning deep neural networks
  for vehicle re-id with visual-spatio-temporal path proposals,'' in
  \emph{ICCV}, 2017.

\bibitem{hermans2017defense}
A.~Hermans, L.~Beyer, and B.~Leibe, ``In defense of the triplet loss for person
  re-identification,'' \emph{arXiv:1703.07737}, 2017.

\bibitem{zheng2017dual}
Z.~Zheng, L.~Zheng, M.~Garrett, Y.~Yang, M.~Xu, and Y.-D. Shen, ``Dual-path
  convolutional image-text embeddings with instance loss,'' \emph{ACM
  Transactions on Multimedia Computing, Communications, and Applications
  (TOMM)}, vol.~16, no.~2, pp. 1--23, 2020.

\bibitem{ding2020adaptive}
Y.~Ding, H.~Fan, M.~Xu, and Y.~Yang, ``Adaptive exploration for unsupervised
  person re-identification,'' \emph{ACM Transactions on Multimedia Computing,
  Communications, and Applications (TOMM)}, vol.~16, no.~1, pp. 1--19, 2020.

\bibitem{zhang2017improving}
Y.~Zhang, D.~Liu, and Z.-J. Zha, ``Improving triplet-wise training of
  convolutional neural network for vehicle re-identification,'' in \emph{ICME},
  2017.

\bibitem{lin2017improving}
Y.~Lin, L.~Zheng, Z.~Zheng, Y.~Wu, Z.~Hu, C.~Yan, and Y.~Yang, ``Improving
  person re-identification by attribute and identity learning,'' \emph{Pattern
  Recognition}, vol.~95, pp. 151--161, 2019.

\bibitem{wang2018transferable}
J.~Wang, X.~Zhu, S.~Gong, and W.~Li, ``Transferable joint attribute-identity
  deep learning for unsupervised person re-identification,'' in \emph{CVPR},
  2018.

\bibitem{wu2019ace}
Z.~Wu, X.~Wang, J.~E. Gonzalez, T.~Goldstein, and L.~S. Davis, ``Ace: Adapting
  to changing environments for semantic segmentation,'' in \emph{ICCV}, 2019.

\bibitem{yue2019domain}
X.~Yue, Y.~Zhang, S.~Zhao, A.~Sangiovanni-Vincentelli, K.~Keutzer, and B.~Gong,
  ``Domain randomization and pyramid consistency: Simulation-to-real
  generalization without accessing target domain data,'' in \emph{ICCV}, 2019.

\bibitem{zheng2017unlabeled}
Z.~Zheng, L.~Zheng, and Y.~Yang, ``Unlabeled samples generated by {GAN} improve
  the person re-identification baseline in vitro,'' in \emph{ICCV}, 2017.

\bibitem{goodfellow2014generative}
I.~Goodfellow, J.~Pouget-Abadie, M.~Mirza, B.~Xu, D.~Warde-Farley, S.~Ozair,
  A.~Courville, and Y.~Bengio, ``Generative adversarial nets,'' in
  \emph{NeurIPS}, 2014.

\bibitem{WahCUB_200_2011}
C.~Wah, S.~Branson, P.~Welinder, P.~Perona, and S.~Belongie, ``The caltech-ucsd
  birds-200-2011 dataset,'' California Institute of Technology, Tech. Rep.
  CNS-TR-2011-001, 2011.

\bibitem{zheng2020university}
Z.~Zheng, Y.~Wei, and Y.~Yang, ``University-1652: A multi-view multi-source
  benchmark for drone-based geo-localization,'' \emph{ACM Multimedia}, 2020.

\bibitem{hoffman2017cycada}
J.~Hoffman, E.~Tzeng, T.~Park, J.-Y. Zhu, P.~Isola, K.~Saenko, A.~A. Efros, and
  T.~Darrell, ``Cycada: Cycle-consistent adversarial domain adaptation,''
  \emph{ICML}, 2018.

\bibitem{zhu2017toward}
J.-Y. Zhu, R.~Zhang, D.~Pathak, T.~Darrell, A.~A. Efros, O.~Wang, and
  E.~Shechtman, ``Toward multimodal image-to-image translation,'' in
  \emph{NeurIPS}, 2017.

\bibitem{tsai2018learning}
Y.-H. Tsai, W.-C. Hung, S.~Schulter, K.~Sohn, M.-H. Yang, and M.~Chandraker,
  ``Learning to adapt structured output space for semantic segmentation,'' in
  \emph{CVPR}, 2018.

\bibitem{tsai2019domain}
Y.-H. Tsai, K.~Sohn, S.~Schulter, and M.~Chandraker, ``Domain adaptation for
  structured output via discriminative patch representations,'' in \emph{ICCV},
  2019.

\bibitem{luo2019taking}
Y.~Luo, L.~Zheng, T.~Guan, J.~Yu, and Y.~Yang, ``Taking a closer look at domain
  shift: Category-level adversaries for semantics consistent domain
  adaptation,'' in \emph{CVPR}, 2019.

\bibitem{zou2018unsupervised}
Y.~Zou, Z.~Yu, V.~Kumar, and J.~Wang, ``Unsupervised domain adaptation for
  semantic segmentation via class-balanced self-training,'' in \emph{ECCV},
  2018.

\bibitem{lee2018diverse}
H.-Y. Lee, H.-Y. Tseng, J.-B. Huang, M.~Singh, and M.-H. Yang, ``Diverse
  image-to-image translation via disentangled representations,'' in
  \emph{ECCV}, 2018.

\bibitem{Zapletal2016}
D.~Zapletal and A.~Herout, ``Vehicle re-identification for automatic video
  traffic surveillance,'' in \emph{CVPR}, 2016.

\bibitem{bai2018group}
Y.~Bai, Y.~Lou, F.~Gao, S.~Wang, Y.~Wu, and L.-Y. Duan, ``Group-sensitive
  triplet embedding for vehicle reidentification,'' \emph{TMM}, vol.~20, no.~9,
  pp. 2385--2399, 2018.

\bibitem{KrauseStarkDengFei-Fei_3DRR2013}
J.~Krause, M.~Stark, J.~Deng, and L.~Fei-Fei, ``3d object representations for
  fine-grained categorization,'' in \emph{3DRR}, 2013.

\bibitem{zheng@joint}
Z.~Zhedong, Y.~Xiaodong, Y.~Zhiding, Z.~Liang, Y.~Yi, and K.~Jan, ``Joint
  discriminative and generative learning for person re-identification,'' in
  \emph{CVPR}, 2019.

\bibitem{russakovsky2015imagenet}
O.~Russakovsky, J.~Deng, H.~Su, J.~Krause, S.~Satheesh, S.~Ma, Z.~Huang,
  A.~Karpathy, A.~Khosla, M.~Bernstein \emph{et~al.}, ``Imagenet large scale
  visual recognition challenge,'' \emph{IJCV}, vol. 115, no.~3, pp. 211--252,
  2015.

\bibitem{he2018deep}
L.~He, J.~Liang, H.~Li, and Z.~Sun, ``Deep spatial feature reconstruction for
  partial person re-identification: Alignment-free approach,'' in \emph{CVPR},
  2018.

\bibitem{huang2017densely}
G.~Huang, Z.~Liu, L.~Van Der~Maaten, and K.~Q. Weinberger, ``Densely connected
  convolutional networks,'' in \emph{CVPR}, 2017.

\bibitem{hu2018senet}
J.~Hu, L.~Shen, and G.~Sun, ``Squeeze-and-excitation networks,'' in
  \emph{CVPR}, 2018.

\bibitem{chen2018virtual}
B.~Chen, W.~Deng, and H.~Shen, ``Virtual class enhanced discriminative
  embedding learning,'' in \emph{NeurIPS}, 2018.

\bibitem{He_2017_ICCV}
K.~He, G.~Gkioxari, P.~Dollar, and R.~Girshick, ``Mask r-cnn,'' in \emph{ICCV},
  2017.

\bibitem{zhong2017re}
Z.~Zhong, L.~Zheng, D.~Cao, and S.~Li, ``Re-ranking person re-identification
  with k-reciprocal encoding,'' in \emph{CVPR}, 2017.

\bibitem{ester1996density}
M.~Ester, H.-P. Kriegel, J.~Sander, X.~Xu \emph{et~al.}, ``A density-based
  algorithm for discovering clusters in large spatial databases with noise.''
  in \emph{KDD}, 1996.

\bibitem{liao2015person}
S.~Liao, Y.~Hu, X.~Zhu, and S.~Li, ``Person re-identification by local maximal
  occurrence representation and metric learning,'' in \emph{CVPR}, 2015.

\bibitem{zhou2017cross}
Y.~Zhou and L.~Shao, ``Cross-view gan based vehicle generation for
  re-identification.'' in \emph{BMVC}, vol.~1, 2017, pp. 1--12.

\bibitem{liu2017provid}
X.~Liu, W.~Liu, T.~Mei, and H.~Ma, ``Provid: Progressive and multimodal vehicle
  reidentification for large-scale urban surveillance,'' \emph{IEEE
  Transactions on Multimedia}, vol.~20, no.~3, pp. 645--658, 2017.

\bibitem{lou2019veri}
Y.~Lou, Y.~Bai, J.~Liu, S.~Wang, and L.~Duan, ``Veri-wild: A large dataset and
  a new method for vehicle re-identification in the wild,'' in \emph{CVPR},
  2019.

\bibitem{zhu2019vehicle}
J.~Zhu, H.~Zeng, J.~Huang, S.~Liao, Z.~Lei, C.~Cai, and L.~Zheng, ``Vehicle
  re-identification using quadruple directional deep learning features,''
  \emph{IEEE Transactions on Intelligent Transportation Systems}, 2019.

\bibitem{khorramshahi2019dual}
P.~Khorramshahi, A.~Kumar, N.~Peri, S.~S. Rambhatla, J.-C. Chen, and
  R.~Chellappa, ``A dual path modelwith adaptive attention for vehicle
  re-identification,'' \emph{arXiv:1905.03397}, 2019.

\bibitem{liu2019pvss}
X.-C. Liu, H.-D. Ma, and S.-Q. Li, ``Pvss: A progressive vehicle search system
  for video surveillance networks,'' \emph{Journal of Computer Science and
  Technology}, vol.~34, no.~3, pp. 634--644, 2019.

\bibitem{chu2019vehicle}
R.~Chu, Y.~Sun, Y.~Li, Z.~Liu, C.~Zhang, and Y.~Wei, ``Vehicle
  re-identification with viewpoint-aware metric learning,'' in \emph{ICCV},
  2019, pp. 8282--8291.

\bibitem{tang2019pamtri}
Z.~Tang, M.~Naphade, S.~Birchfield, J.~Tremblay, W.~Hodge, R.~Kumar, S.~Wang,
  and X.~Yang, ``Pamtri: Pose-aware multi-task learning for vehicle
  re-identification using highly randomized synthetic data,'' in \emph{ICCV},
  2019.

\bibitem{qian2019stripe}
J.~Qian, W.~Jiang, H.~Luo, and H.~Yu, ``Stripe-based and attribute-aware
  network: A two-branch deep model for vehicle re-identification,''
  \emph{arXiv:1910.05549}, 2019.

\bibitem{he2019part}
B.~He, J.~Li, Y.~Zhao, and Y.~Tian, ``Part-regularized near-duplicate vehicle
  re-identification,'' in \emph{CVPR}, 2019.

\bibitem{he2016deep}
K.~He, X.~Zhang, S.~Ren, and J.~Sun, ``Deep residual learning for image
  recognition,'' in \emph{CVPR}, 2016.

\bibitem{tan2019multi}
X.~Tan, Z.~Wang, M.~Jiang, X.~Yang, J.~Wang, Y.~Gao, X.~Su, X.~Ye, Y.~Yuan,
  D.~He \emph{et~al.}, ``Multi-camera vehicle tracking and re-identification
  based on visual and spatial-temporal features,'' in \emph{CVPR Workshops},
  2019.

\bibitem{huang2019multi}
T.-W. Huang, J.~Cai, H.~Yang, H.-M. Hsu, and J.-N. Hwang, ``Multi-view vehicle
  re-identification using temporal attention model and metadata re-ranking,''
  in \emph{CVPR Workshops}, 2019.

\bibitem{lv2019vehicle}
K.~Lv, W.~Deng, Y.~Hou, H.~Du, H.~Sheng, J.~Jiao, and L.~Zheng, ``Vehicle
  reidentification with the location and time stamp,'' in \emph{CVPR
  Workshops}, 2019.

\bibitem{zheng2016discriminatively}
Z.~Zheng, L.~Zheng, and Y.~Yang, ``A discriminatively learned cnn embedding for
  person reidentification,'' \emph{ACM Transactions on Multimedia Computing,
  Communications, and Applications (TOMM)}, vol.~14, no.~1, pp. 1--20, 2017.

\bibitem{yao2019simulating}
Y.~Yao, L.~Zheng, X.~Yang, M.~Naphade, and T.~Gedeon, ``Simulating content
  consistent vehicle datasets with attribute descent,''
  \emph{arXiv:1912.08855}, 2019.

\bibitem{zhang2019leveraging}
D.~Zhang, J.~Han, L.~Zhao, and D.~Meng, ``Leveraging prior-knowledge for weakly
  supervised object detection under a collaborative self-paced curriculum
  learning framework,'' \emph{International Journal of Computer Vision}, vol.
  127, no.~4, pp. 363--380, 2019.

\bibitem{zhang2018spftn}
D.~Zhang, J.~Han, L.~Yang, and D.~Xu, ``Spftn: a joint learning framework for
  localizing and segmenting objects in weakly labeled videos,'' \emph{IEEE
  transactions on pattern analysis and machine intelligence}, 2018.

\bibitem{meng2019weakly}
J.~Meng, S.~Wu, and W.-S. Zheng, ``Weakly supervised person
  re-identification,'' in \emph{CVPR}, 2019, pp. 760--769.

\end{thebibliography}

\begin{IEEEbiography}
[{\includegraphics[width=1in,height=1.2in,clip,keepaspectratio]{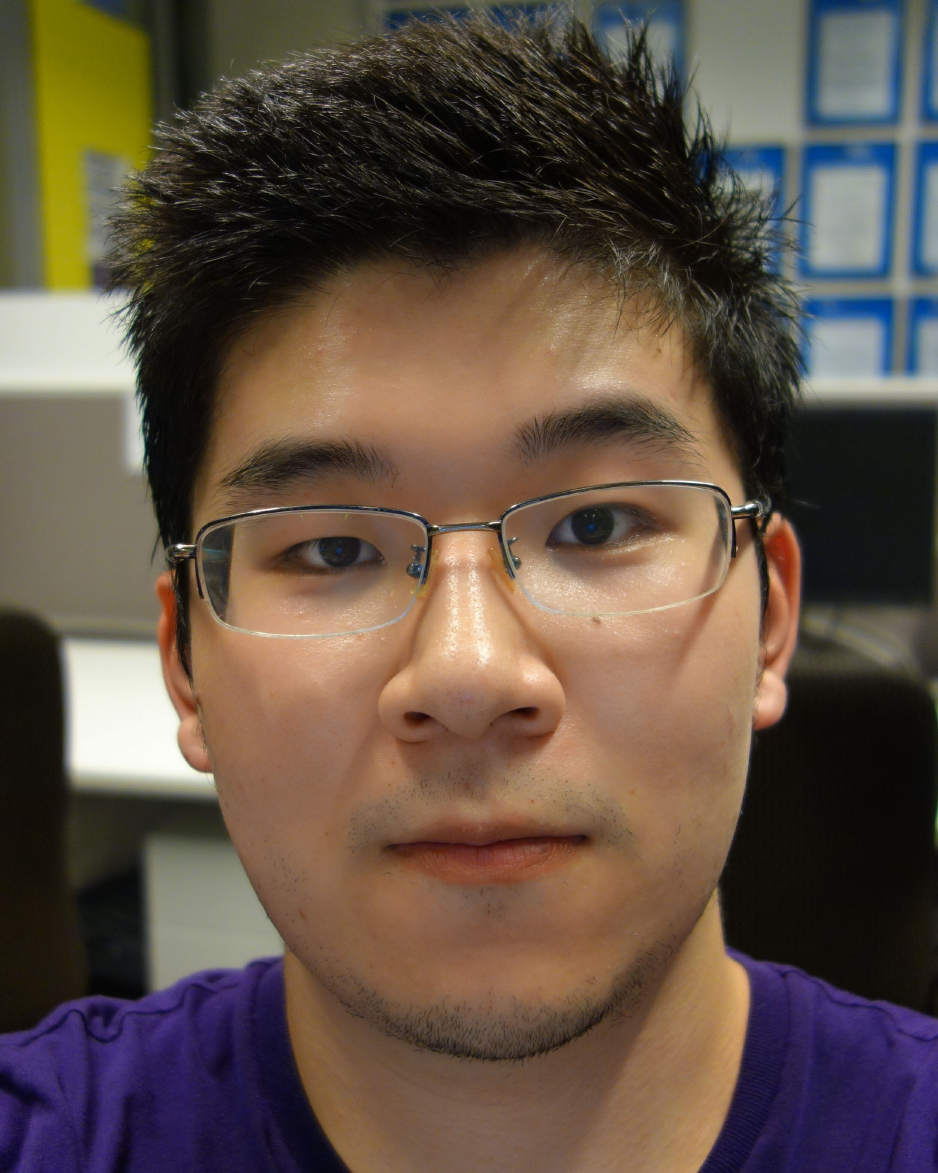}}]{Zhedong Zheng}
received the B.S. degree in computer science from Fudan University, China, in 2016. He is currently a Ph.D. student with the School of Computer Science at University of Technology Sydney, Australia. His research interests include robust learning for image retrieval, generative learning for data augmentation, and unsupervised domain adaptation.
\end{IEEEbiography}

\vfill
\begin{IEEEbiography}
[{\includegraphics[width=1in,height=1.2in,clip,keepaspectratio]{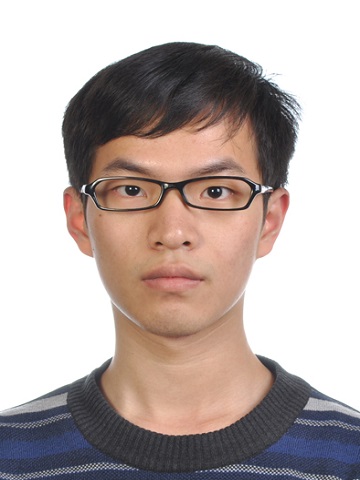}}]{Tao Ruan} received the B.E. degree in computer science and technology from Beijing Jiaotong University, Beijing, China, in 2016. He is currently a Ph.D. candidate with the Institute of Information Science, Beijing Jiaotong University, Beijing, China. His research interests include video synopsis and pixel understanding.
\end{IEEEbiography}

\vfill
\begin{IEEEbiography}
[{\includegraphics[width=1in,height=1.2in,clip,keepaspectratio]{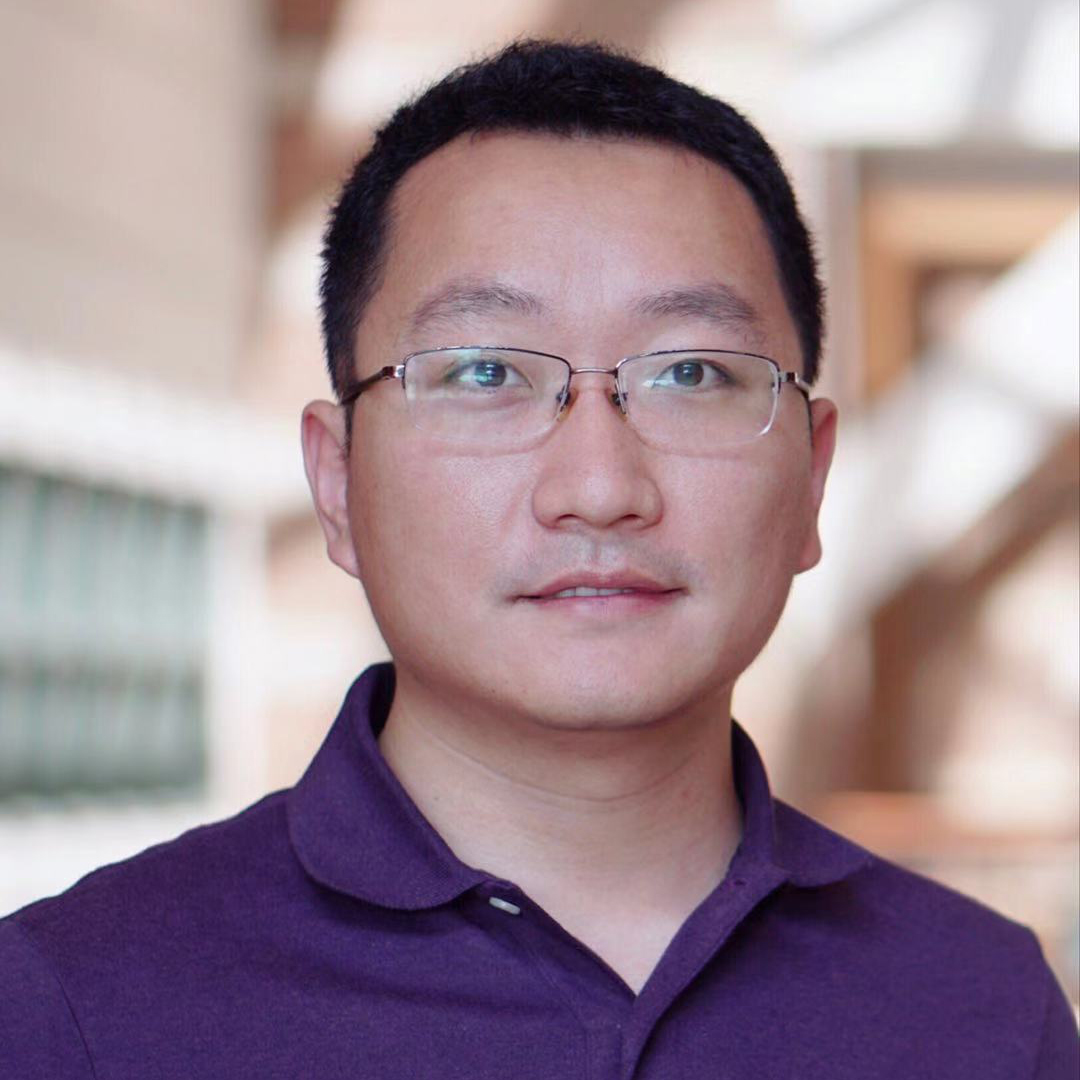}}]{Yunchao Wei} received his Ph.D. degree from Beijing Jiaotong University, Beijing, China. He was a Postdoctoral Researcher at Beckman Institute, UIUC, from 2017 to 2019. He is currently an Assistant Professor with the Australian Artificial Intelligence Institute (AAII), University of Technology Sydney. He is ARC Discovery Early Career Researcher Award (DECRA) Fellow from 2019 to 2021. 
His current research interests include computer vision and machine learning.
\end{IEEEbiography}

\vfill
\begin{IEEEbiography}[{\includegraphics[width=1in,height=1.2in,clip,keepaspectratio]{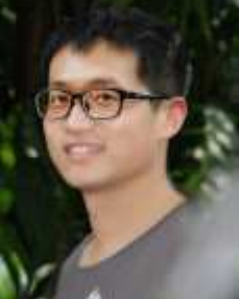}}]{Yi Yang} received the Ph.D. degree in computer
science from Zhejiang University, Hangzhou, China, in 2010. He is currently a professor with University of Technology Sydney, Australia.
He was a Post-Doctoral Research with the School of Computer Science, Carnegie Mellon University, Pittsburgh, PA, USA. His current research interest includes machine learning and its applications to multimedia content analysis and computer vision. %, such as multimedia indexing and retrieval, video analysis and video semantics understanding.
\end{IEEEbiography}

\vfill
\begin{IEEEbiography}[{\includegraphics[width=1in,height=1.2in,clip,keepaspectratio]{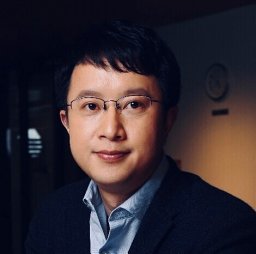}}]{Tao Mei}  is a Technical Vice
President with JD.COM and the Deputy Managing
Director of AI Research of JD.COM, where he also serves
as the Director of Computer Vision and Multimedia
Lab. Prior to joining JD.COM in 2018, he was a
Senior Research Manager with Microsoft Research
Asia in Beijing, China. He has authored or coauthored over 200 publications (with 12 best paper
awards) in journals and conferences, 10 book chapters, and edited five books. He holds over 50 US and
international patents (20 granted). He is or has been
an Editorial Board Member of IEEE Trans. on Image Processing, IEEE Trans.
on Circuits and Systems for Video Technology, IEEE Trans. on Multimedia,
ACM Trans. on Multimedia, Pattern Recognition, etc. Tao received B.E. and
Ph.D. degrees from the University of Science and Technology of China,
Hefei, China, in 2001 and 2006, respectively. He was elected as a Fellow of
IEEE (2019), a Fellow of IAPR (2016), a Distinguished Scientist of ACM
(2016), and a Distinguished Industry Speaker of IEEE Signal Processing
Society (2017), for his contributions to large-scale multimedia analysis and
applications.
%\begin{IEEEbiography}[{\includegraphics[width=1in,height=1.25in,clip,keepaspectratio]{image/zhuang.jpeg}}]{Yueting Zhuang} received the B.Sc., M.Sc.,and Ph.D. degrees in computer science from Zhejiang University, China, in 1986, 1989, and 1998, respectively. From February 1997 to August 1998, he was a Visiting Scholar with the University of Illinois at Urbana-Champaign. He has served as the Dean of the College of Computer Science, Zhejiang University, from 2009 to 2017, the Director of the Institute of Artificial Intelligence, from 2006 to 2015. He is currently a Full Professor with the College of Computer Science, the Director of the MOE-Digital Library Engineering Research Center, Zhejiang University. His research interests mainly include multimedia retrieval, artificial intelligence, cross-media computing, and digital library.
\end{IEEEbiography}

\end{document}